\documentclass{article}
\usepackage{float}
\usepackage{colm2024_conference}

\usepackage{booktabs}
\usepackage{graphicx}
\usepackage{tabularx}
\usepackage{array}
\newcolumntype{L}{>{\raggedright\arraybackslash}X}
\newcolumntype{P}[1]{>{\raggedright\arraybackslash}p{#1}}
\usepackage{enumitem}
\usepackage{wrapfig}
\usepackage{algorithm}
\usepackage{algpseudocode}
\usepackage{natbib}
\usepackage{makecell}
\usepackage{bbm}
\usepackage{amsmath}
\usepackage{amssymb}
\usepackage{amsfonts}
\usepackage{multirow}
\usepackage{verbatim}
\usepackage{caption}
\usepackage{longtable}
\usepackage{supertabular}

\usepackage{CJKutf8}
\usepackage[utf8]{inputenc}
\usepackage[T1]{fontenc}
\usepackage{microtype}
\usepackage{marvosym}
\usepackage{pifont}
\usepackage{afterpage}
\usepackage{tablefootnote}
\usepackage{xspace}
\usepackage{textcomp}
\usepackage{siunitx}
\usepackage{xcolor}
\usepackage{colortbl}
\usepackage{adjustbox}

\definecolor{dt}{gray}{0.7}
\definecolor{tabhl}{RGB}{227,227,250}
\definecolor{tongyi-purple}{RGB}{97,92,237}
\colorlet{tongyi-purple-alpha}{tongyi-purple!38}

\newcommand{\cmark}{\ding{51}}%
\newcommand{\xmark}{\ding{55}}%

\usepackage{tgpagella}
\usepackage{latexsym}
\definecolor{mydarkblue}{rgb}{0,0.08,0.45}
\definecolor{citecolor}{HTML}{0071BC}
\usepackage{url}
\usepackage{nicefrac}
\usepackage{changepage}
\usepackage{subcaption}
\usepackage{multicol}
\usepackage{titlesec}
\titleformat*{\section}{\large\bfseries}

\usepackage{hyperref}
\usepackage{cleveref}

\newcommand{\ours}{Qwen-Drive-1.0\xspace}

\title{\ours{}: An Initial Step towards a Vision-Language Foundation Model for Autonomous Driving}

\author{
\bf Qwen Team \\ Huazhong University of Science and Technology
}

\begin{document}

\maketitle

\begin{abstract}
We present \ours{}, an initial step towards a vision-language foundation model for autonomous driving. \ours{} retains the architecture of the pretrained vision-language model (VLM) and integrates 3D perception, visual question answering, and motion planning within a unified framework. An external bird's-eye-view (BEV) perception head jointly performs 3D object detection, semantic occupancy prediction, and BEV map segmentation. It serves as a probe of the 3D information accessible from the shared representations and provides an explicit, inspectable interface to 3D scene structure. A Planning Expert conditions on shared VLM representations to generate future ego trajectories. A staged training recipe combines driving supervision with general-purpose vision-language data to acquire driving-specific competence while helping preserve broad visual understanding and instruction-following capabilities. Experiments demonstrate strong 3D perception and driving scene understanding while largely preserving general vision-language capability. Comprehensive evaluations across open-loop, pseudo-closed-loop, and closed-loop settings further show highly competitive motion-planning performance.
\end{abstract}

\begin{figure*}[h]
\vspace{2em}
\centering
\includegraphics[width=\textwidth]{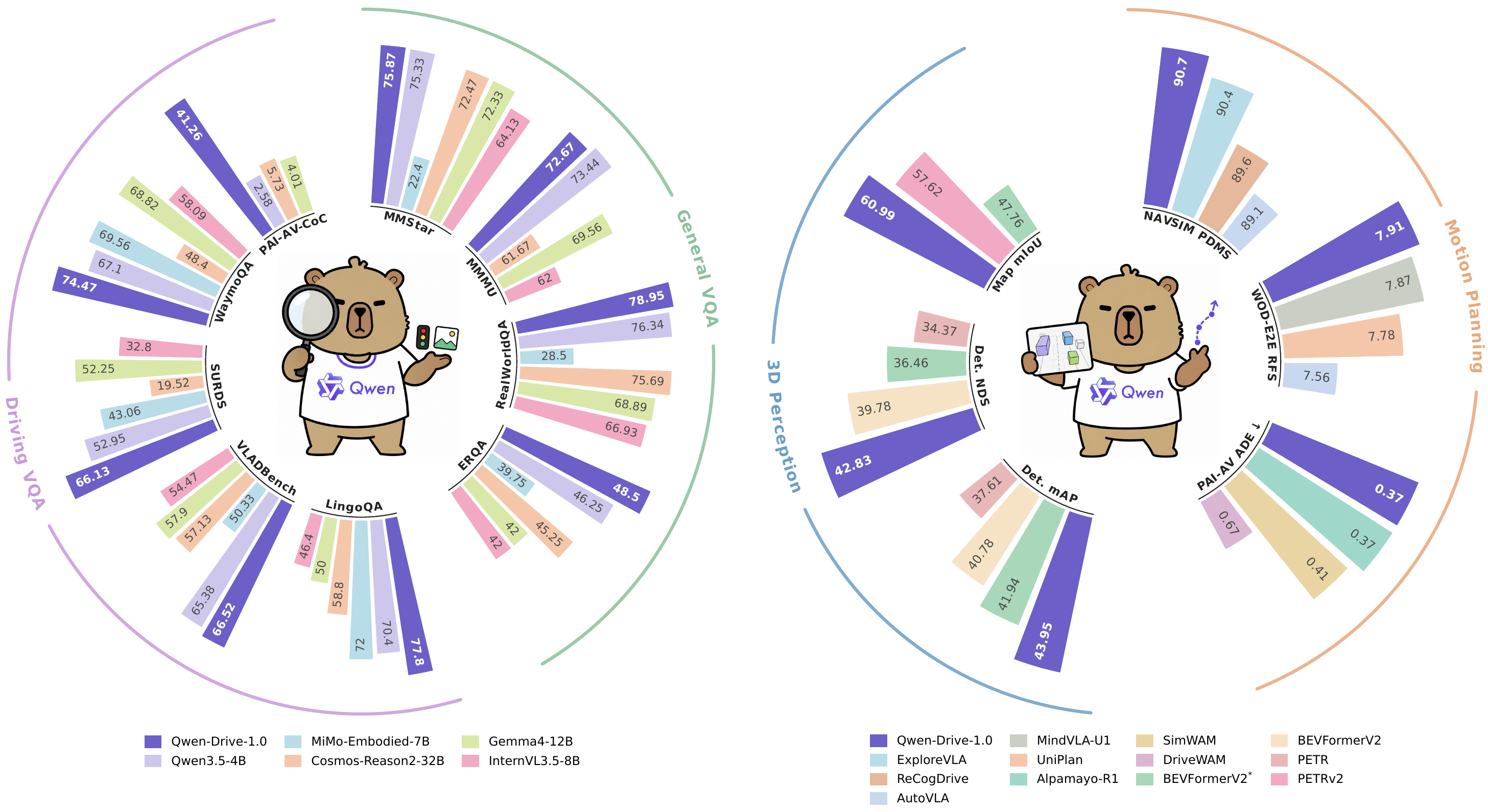}
\caption{
    Performance overview of \ours{} across driving VQA, general VQA, 3D perception, and motion planning.
}
\label{fig:intro}
\end{figure*}
\newpage

\section{Introduction}

Autonomous driving research has increasingly shifted from task-specific modular pipelines towards unified learning-based systems~\citep{hu2023planning}. Within this trend, vision-language-action (VLA) models use pretrained vision-language models (VLMs) to connect scene understanding, reasoning, and action generation~\citep{fu2025orion,fu2025minddrive,wang2026qwen}. Large-scale pretraining provides broad visual, linguistic, and world knowledge that can support reasoning in rare and out-of-distribution (OOD) driving scenarios.

Many recent driving VLA methods adapt a general-purpose VLM through continued training on driving-specific supervision, particularly visual question answering (VQA). This recipe expresses heterogeneous driving tasks through a common autoregressive language interface. The resulting targets cover traffic-scene description, reasoning about surrounding agents, and explanations of driving decisions~\citep{sima2024drivelm,xu2024drivegpt4,zhou2025hermes,wang2025omnidrive,zhao2026extending}. Driving-specific knowledge is therefore introduced primarily through language supervision.

This adaptation strategy has two limitations. \textbf{1)} Textual VQA targets do not directly constrain 3D layout, depth, or occupancy~\citep{yang2025thinking,elbanani2024probing,zhou2026hermes++,wu2026generation}. Even when pretrained representations encode spatial cues, textual supervision alone neither requires explicit 3D predictions nor permits their direct evaluation. A model adapted only through VQA can therefore produce fluent scene descriptions while remaining imprecise in 3D space. \textbf{2)} Extensive domain adaptation can cause catastrophic forgetting of the general knowledge acquired during pretraining~\citep{zhai2024investigating,luo2025empirical}. No finite driving dataset can exhaustively represent the rare and unseen situations encountered in deployment. This pretrained knowledge therefore remains important for OOD reasoning. Together, these limitations motivate a unified model for 3D perception, driving reasoning, and motion planning that retains broad visual and world knowledge from pretraining.

Preserving general capability is also a deployment requirement. Production vehicles are moving towards cockpit-driving integration, in which the intelligent cockpit and the driving system share a single compute platform rather than two separate domain controllers. This consolidation lowers hardware and integration costs, and it tightens the compute budget available to each function. A single model is then expected to serve both domains, which requires general capabilities such as multi-turn dialogue, instruction following, and open-ended visual understanding in addition to driving competence. A model that trades general capability for driving performance forfeits this benefit, because the cockpit functions would then require a separate model and additional compute. Retaining general capability therefore serves two purposes. It supports reasoning in rare and unseen situations, and it allows one model to cover both the cockpit and the driving domain within a single compute budget.

We argue that a practical vision-language foundation model for driving should satisfy three design requirements. First, the pretrained VLM architecture should remain unchanged to preserve ease of use. Second, an explicit perception probe should expose and evaluate 3D scene information rather than relying on textual spatial reasoning alone. Third, the model should acquire driving scene-understanding knowledge while retaining most of its general-purpose capabilities, which supports both robust generalization and deployment on an integrated cockpit-driving platform.

We therefore introduce \ours{}, the first vision-language foundation model for autonomous driving to our knowledge that unifies 3D perception, visual question answering, and motion planning within a single pretrained VLM. \ours{} uses the natively multimodal Qwen3.5-4B~\citep{qwen35blog,qwen35modelcard} as the shared VLM and attaches two external modules. The bird's-eye-view (BEV) perception head probes the shared representations through explicit and inspectable 3D scene predictions. A Planning Expert uses these representations to generate future ego trajectories. We train these components with a staged recipe that introduces perception, language, and planning objectives. A unified data pipeline underpins this recipe, mapping heterogeneous perception annotations into a shared label space, re-annotating driving VQA responses for format and factual consistency, and expressing trajectories from multiple public driving datasets in a single waypoint representation.

Fig.~\ref{fig:intro} summarizes the performance of \ours{} across 3D perception, driving scene understanding, general vision-language capabilities, and motion planning. On 3D perception, it reaches 43.95 mAP and 60.99 map mIoU on nuScenes and 43.45 mAP and 71.27 map mIoU on OpenScene, remaining highly competitive with common vision-based 3D detectors and demonstrating the explicit 3D perception capability added to the pretrained VLM. On driving scene understanding, it significantly surpasses the general-purpose Qwen3.5-4B while preserving general capability. On motion planning, it achieves a Predictive Driver Model Score of 90.7 on NAVSIM, attains a strong Rater Feedback Score of 7.91 on the test split of the Waymo Open Dataset end-to-end benchmark (WOD-E2E), and shows promising potential for closed-loop driving in AlpaSim.

We summarize our main contributions below.

\begin{itemize}
    \item We present \ours{}, to our knowledge the first vision-language foundation model for autonomous driving that integrates 3D perception, driving VQA, and motion planning without changing the pretrained VLM architecture.

    \item We introduce an external BEV perception head that jointly learns 3D detection, semantic occupancy prediction, and BEV map segmentation. The head serves as a 3D probe and equips the same pretrained VLM with explicit, inspectable perception outputs while preserving highly competitive vision-language performance.

    \item We develop a staged training and data recipe that unifies cross-dataset labels, rewrites responses, filters samples for consistency, and combines driving data with general-purpose vision-language supervision. This design supports domain adaptation while mitigating catastrophic forgetting.
    
    \item We design a Planning Expert tailored to pretrained VLM representations, using flow matching to generate future ego trajectories. Unified trajectory annotations enable joint training across multiple public driving datasets and yield highly competitive results across open-loop, pseudo-closed-loop, and closed-loop evaluations.
    
\end{itemize}

\section{Method}
\label{sec:method}

\begin{figure*}[t]
\centering
\includegraphics[width=\textwidth]{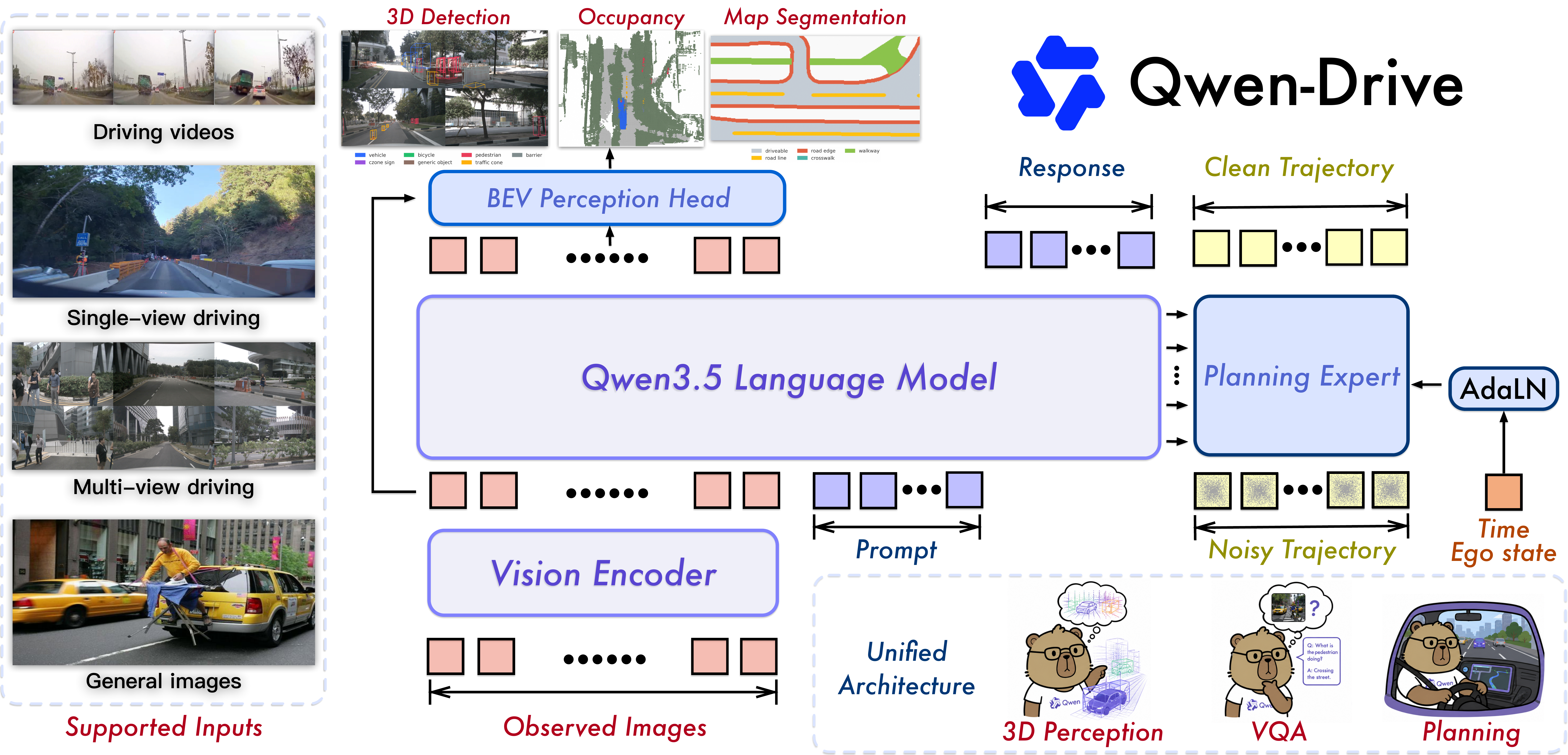}
\caption{
    Unified architecture of \ours{} for 3D perception, visual question answering, and motion planning. A shared vision encoder and VLM support text generation, while the external BEV perception head and Planning Expert produce geometric predictions and future ego trajectories.
}
\label{fig:arch}
\end{figure*}

\subsection{Model Architecture and Objectives}
\label{sec:arch}

Fig.~\ref{fig:arch} presents the unified architecture of \ours{}. A shared vision encoder and VLM process single-view and multi-view driving inputs, temporal image sequences, and general images. The vision encoder converts each image into visual tokens. The VLM encodes these tokens with the textual prompt and generates responses autoregressively. Two external modules use features from this shared pathway without changing the VLM architecture. The BEV perception head fuses vision encoder features with VLM output features to construct a BEV representation for 3D object detection, semantic occupancy prediction, and BEV map segmentation. The Planning Expert conditions trajectory tokens on cached VLM keys and values and predicts future ego motion through flow matching.

\paragraph{Multi-View and Multi-Frame Inputs.}
A visual token sequence does not explicitly identify the view and timestep of each image, so we provide this information through view and frame tags. The view tags denote eight canonical directions, namely \texttt{<FRONT VIEW>}, \texttt{<FRONT RIGHT VIEW>}, \texttt{<RIGHT VIEW>}, \texttt{<BACK RIGHT VIEW>}, \texttt{<BACK VIEW>}, \texttt{<BACK LEFT VIEW>}, \texttt{<LEFT VIEW>}, and \texttt{<FRONT LEFT VIEW>}. The frame tag \texttt{frame: $k$} associates each image with timestep $k$.

Input serialization depends on the task. Question answering examples use frame-major order, which places all views at one timestep before those at the next timestep:
\[
\texttt{frame: 0 <FRONT VIEW> <image> <FRONT RIGHT VIEW> <image>}\,\cdots\,\texttt{frame: 1}\,\cdots\, .
\]
Planning examples, including our self-constructed planning-reasoning data, use view-major order:
\[
\texttt{<FRONT VIEW> frame: 0 <image> frame: 1 <image>}\,\cdots\,\texttt{<FRONT RIGHT VIEW>}\,\cdots\, .
\]
View-major serialization places consecutive observations from each view adjacent in the token sequence and exposes temporal variation within that view, which is important for control in dynamic environments~\citep{fang2026towards}. Single-view and single-frame inputs omit the corresponding redundant tags. Both tag types use ordinary vocabulary tokens and require no additional special tokens or architectural modifications.

\paragraph{Autoregressive Text Generation.}
Given a serialized multimodal input $\mathbf{x}$ and a target response $\mathbf{y}=(y_1,\ldots,y_T)$, the VLM predicts each response token conditioned on the input and preceding tokens. We use the standard next-token prediction objective:
\begin{equation}
\mathcal{L}_{\mathrm{ntp}}
=
-\sum_{t=1}^{T}
\log p\!\left(y_t \mid \mathbf{x},y_{<t}\right).
\label{eq:ntploss}
\end{equation}
We use the same objective for driving-specific and general-purpose vision-language samples.

\begin{figure*}[t]
\centering
\includegraphics[width=\textwidth]{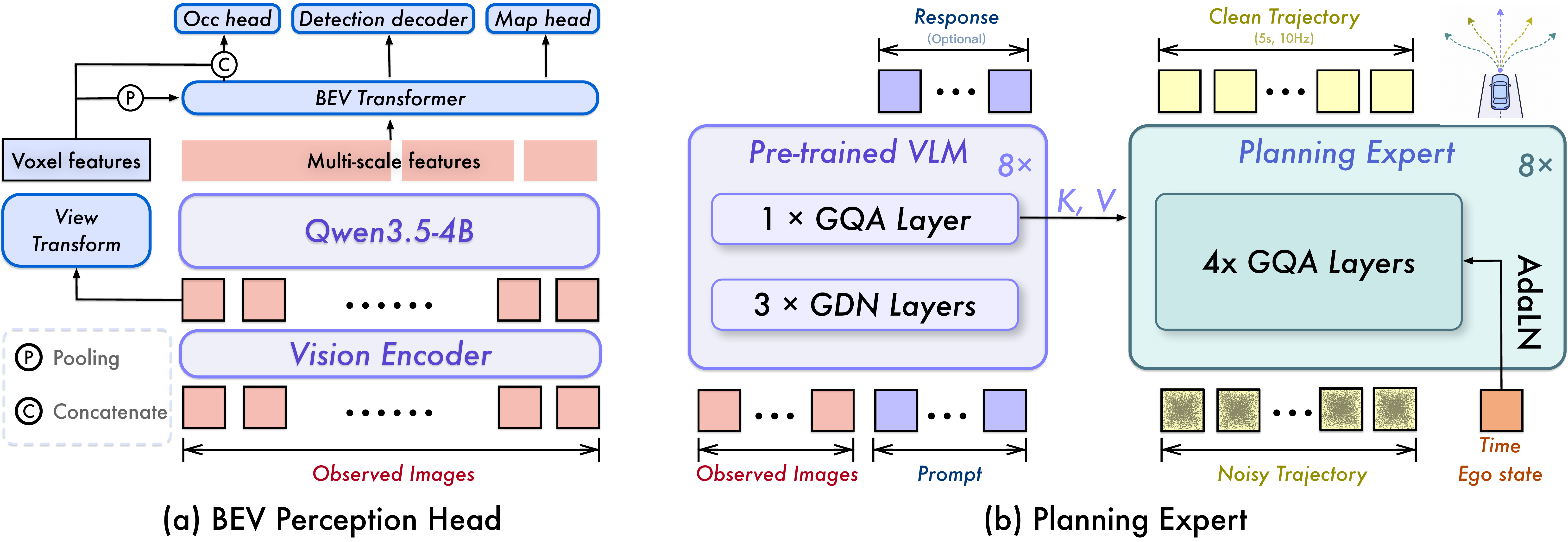}
\caption{
    Architectures of the external modules. (a) The BEV perception head fuses voxelized vision encoder features with a feature pyramid of VLM outputs. (b) The Planning Expert conditions noisy trajectory tokens on cached VLM keys and values to recover a clean ego trajectory.
}
\label{fig:head}
\end{figure*}

\paragraph{BEV Perception Head.}
As shown in Fig.~\ref{fig:head}(a), the BEV perception head performs single-frame surround-view 3D perception. It receives $N_v$ current images and their camera calibrations, where $N_v\in\{6,8\}$ in our experiments. The head constructs a shared ego-frame BEV representation for 3D object detection, semantic occupancy prediction, and BEV map segmentation.

The BEV perception head reads two complementary feature streams from each view $i$. The vision encoder feature $\mathbf{F}^{v}_{i}$ captures low-level appearance before the image tokens enter the VLM. After traversing the full VLM, the corresponding image tokens yield the feature $\mathbf{F}^{m}_{i}$, which encodes broader scene context and serves as the semantic source for BEV construction. We denote the feature sequences across views by $\mathbf{F}^{v}=(\mathbf{F}^{v}_{i})_{i=1}^{N_v}$ and $\mathbf{F}^{m}=(\mathbf{F}^{m}_{i})_{i=1}^{N_v}$. During joint training, the perception losses propagate through $\mathbf{F}^{m}_{i}$, providing an additional gradient path to the vision encoder alongside the direct path through $\mathbf{F}^{v}_{i}$.

To construct an explicit geometric representation, a depth-based view transform~\citep{li2022unifying,philion2020lift} lifts the single-scale features $\mathbf{F}^{v}$ into a 3D volume without a feature pyramid. A lightweight depth network comprising residual blocks and an atrous spatial pyramid predicts a per-pixel categorical distribution $\mathbf{D}_{i}$ over $N_d$ depth bins without depth supervision. Each voxel center $\mathbf{p}$ within the perception range is projected into view $i$ using the calibration matrix $\mathbf{P}_i$, yielding image coordinates $(u_i,v_i)$ and depth bin $d_i$. Its voxel feature is computed as:
\begin{equation}
\mathbf{V}(\mathbf{p}) = \sum_{i \in \Omega(\mathbf{p})} \mathbf{D}_{i}(u_i,v_i,d_i)\,\mathbf{F}^{v}_{i}(u_i,v_i),
\label{eq:lift}
\end{equation}
where $\Omega(\mathbf{p})$ contains the views in which $\mathbf{p}$ has a valid image projection. This operation distributes image features along camera rays according to the predicted depth probabilities. The resulting volume $\mathbf{V}$ retains the height dimension for occupancy prediction.

Because $\mathbf{F}^{m}$ is available at a single coarse scale, a simple feature pyramid~\citep{li2022exploring} expands it into multiscale features. A query-based BEV transformer~\citep{li2022bevformer,yang2023bevformer} then aggregates these features onto the BEV plane. Its queries are initialized using the height-collapsed feature $\bar{\mathbf{V}}$ derived from $\mathbf{V}$, which provides an explicit geometric prior. Each encoder layer alternates self-attention over the BEV grid with deformable cross-attention to the feature pyramid. The resulting ego-frame feature $\mathbf{B}$ integrates geometry from $\mathbf{V}$ with context from $\mathbf{F}^{m}$ and serves all three task-specific branches.

For 3D detection, a DETR-style decoder with deformable attention~\citep{zhu2020deformable} refines object queries against $\mathbf{B}$. For semantic occupancy, we expand $\mathbf{B}$ along the height dimension and fuse it with $\mathbf{V}$ before a shallow 3D UNet predicts per-voxel semantics. This fusion restores the vertical structure retained in $\mathbf{V}$. For map segmentation, a UNet-style head predicts rasterized map elements on the BEV plane. We jointly optimize the three branches using the perception objective:
\begin{equation}
\mathcal{L}_{\mathrm{perc}}
=
\mathcal{L}_{\mathrm{det}}
+
\mathcal{L}_{\mathrm{occ}}
+
\mathcal{L}_{\mathrm{map}}.
\label{eq:percloss}
\end{equation}

Detection follows the set-prediction formulation. The Hungarian algorithm matches object queries to ground-truth boxes, and deep supervision at each decoder layer combines a focal loss~\citep{lin2017focal} with an $\ell_1$ regression loss:
\begin{equation}
\mathcal{L}_{\mathrm{det}} = \sum_{l=1}^{L} \Big( 2\, \mathcal{L}^{(l)}_{\mathrm{focal}} + 0.75\, \mathcal{L}^{(l)}_{\ell_1} \Big).
\label{eq:detloss}
\end{equation}
Following FlashOcc~\citep{yu2023flashocc}, the occupancy objective is defined as:
\begin{equation}
\mathcal{L}_{\mathrm{occ}} = 100\, \mathcal{L}_{\mathrm{focal}} + \mathcal{L}_{\mathrm{geo}} + \mathcal{L}_{\mathrm{sem}} + \mathcal{L}_{\mathrm{lov}},
\label{eq:occloss}
\end{equation}
where $\mathcal{L}_{\mathrm{focal}}$ is a class-balanced focal loss, $\mathcal{L}_{\mathrm{geo}}$ and $\mathcal{L}_{\mathrm{sem}}$ are the geometric and semantic scene-class affinity losses of MonoScene~\citep{cao2022monoscene}, and $\mathcal{L}_{\mathrm{lov}}$ is the Lov\'asz-softmax loss~\citep{berman2018lovasz}. The map objective is $\mathcal{L}_{\mathrm{map}}=100\,\mathcal{L}_{\mathrm{focal}}+\mathcal{L}_{\mathrm{lov}}$.

\paragraph{Planning Expert.}
For motion planning, the Planning Expert predicts future ego motion from the multimodal context encoded by the VLM. We formulate trajectory prediction as conditional generation:
\begin{equation}
\boldsymbol{\tau} \sim p\!\left(
\boldsymbol{\tau} \mid
\mathbf{s}, \ell, \boldsymbol{\tau}_{\mathrm{hist}}, \mathbf{n}, \mathbf{e}, \mathbf{r}
\right),
\qquad
\boldsymbol{\tau} = \{(x_k, y_k, \theta_k)\}_{k=1}^{50},
\label{eq:traj}
\end{equation}
where $\mathbf{s}$ denotes the vehicle sensor inputs and $\ell$ denotes their serialized layout. The variables $\boldsymbol{\tau}_{\mathrm{hist}}$, $\mathbf{n}$, and $\mathbf{e}$ denote the historical ego trajectory, navigation instruction, and current ego state, respectively. The optional textual planning reason $\mathbf{r}$ is set to $\varnothing$ when unavailable. The prompt describes $\ell$, provides $\boldsymbol{\tau}_{\mathrm{hist}}$ and $\mathbf{n}$, and includes $\mathbf{r}$ when available. Each trajectory contains 50 waypoints spanning 5\,s at 10\,Hz. At waypoint $k$, $x_k$ and $y_k$ denote the longitudinal and lateral positions in the current ego frame, while $\theta_k$ denotes the heading relative to the current ego orientation. For joint training across datasets, we divide $x_k$, $y_k$, and $\theta_k$ by fixed scales of 165\,m, 25\,m, and $\pi/2$\,rad, respectively.

As illustrated in Fig.~\ref{fig:head}(b), the Planning Expert uses a 32-layer diffusion transformer. The VLM alternates gated linear attention with grouped-query softmax attention~\citep{qwen35modelcard}. We cache the keys after rotary position embedding (RoPE) and the corresponding values from all eight grouped-query softmax attention layers. Each cache conditions four consecutive Planning Expert layers. Each Planning Expert layer concatenates the cached keys and values with those of the trajectory tokens for joint attention. A trajectory token combines a noisy waypoint with an encoding of $\boldsymbol{\tau}_{\mathrm{hist}}$. Shared adaptive layer normalization injects the flow time, navigation instruction $\mathbf{n}$, and current ego state $\mathbf{e}$. The hidden dimension is 1024, yielding $\sim$1.1B parameters.

We train the Planning Expert by flow matching~\citep{lipman2023flow} with an $x$-prediction parameterization that directly estimates the clean trajectory. Let $\boldsymbol{\tau}_1$ denote the normalized ground-truth trajectory, and let $\boldsymbol{\tau}_0 \sim \mathcal{N}(\mathbf{0}, \mathbf{I})$ denote a Gaussian noise sample of the same shape. We define the linear interpolation path at flow time $t$ as:
\begin{equation}
\boldsymbol{\tau}_t = (1 - t)\, \boldsymbol{\tau}_0 + t\, \boldsymbol{\tau}_1.
\label{eq:flowinterp}
\end{equation}
The Planning Expert predicts the clean endpoint $\hat{\boldsymbol{\tau}}_1$ rather than the flow velocity or noise. The predicted endpoint induces the flow velocity field $(\hat{\boldsymbol{\tau}}_1-\boldsymbol{\tau}_t)/(1-t)$. This endpoint parameterization reduces sensitivity to sensor noise in trajectories recorded across heterogeneous datasets. To keep this conversion well conditioned, we sample $\tilde{t} \sim \mathrm{Beta}(1.5, 1.0)$ and set $t = \min\{\tilde{t}, 0.9\}$, ensuring $1-t \geq 0.1$. The complete objective combines flow matching with temporal regularization:
\begin{equation}
\mathcal{L}_{\mathrm{plan}} = \mathcal{L}_{\mathrm{fm}} + 2\times10^{-4}\, \mathcal{L}_{\Delta^{1}} + 2\times10^{-5}\, \mathcal{L}_{\Delta^{2}},
\label{eq:planloss}
\end{equation}
where $\mathcal{L}_{\mathrm{fm}}$ is the squared error between the induced flow velocity and the target flow velocity $\boldsymbol{\tau}_1-\boldsymbol{\tau}_0$. For $j \in \{1,2\}$, $\mathcal{L}_{\Delta^{j}}$ is a Huber penalty that matches the $j$th-order temporal differences of $\hat{\boldsymbol{\tau}}_1$ and $\boldsymbol{\tau}_1$. Together, these temporal regularizers discourage waypoint jitter and abrupt changes in acceleration.

At inference, Gaussian noise initializes the trajectory tokens. A 10-step Euler solver then integrates the induced flow velocity field to obtain the final trajectory.

\subsection{Training Recipe}
\label{sec:training}

\begin{figure*}[t]
\centering
\includegraphics[width=\textwidth]{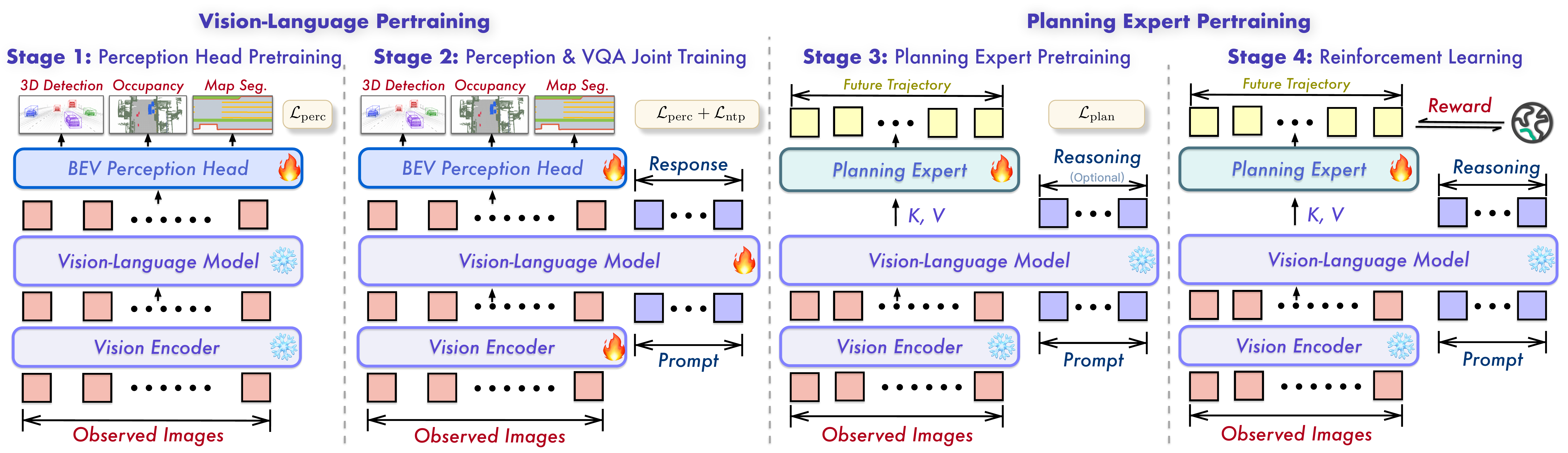}
\caption{
    Four-stage training recipe of \ours{}. Stages 1 and 2 adapt the shared vision-language pathway, first initializing the BEV perception head and then using perception and VQA supervision to update the vision encoder and VLM. Stages 3 and 4 train the Planning Expert on top of these fixed representations, first by flow matching and then by reward-based optimization. Flames indicate trainable modules, and snowflakes indicate fixed modules.
}
\label{fig:training_recipe}
\end{figure*}

As illustrated in Fig.~\ref{fig:training_recipe}, we train \ours{} in four stages. The first two stages initialize the BEV perception head and jointly adapt the shared pathway for explicit 3D prediction. Stage 3 trains trajectory generation with optional textual reasoning as a condition. Stage 4 further refines the resulting model through reinforcement-based optimization.

\paragraph{Stage 1. Perception Head Pretraining.}
We keep the vision encoder and VLM fixed and optimize only the newly initialized BEV perception head with $\mathcal{L}_{\mathrm{perc}}$. Its view transform, BEV transformer, and task decoders learn to construct and decode an ego-frame representation. This stage initializes the newly added module before joint adaptation in Stage 2.

\paragraph{Stage 2. Perception and VQA Joint Training.}
Sec.~\ref{sec:exp_perception} will show that head-only training yields limited perception performance, indicating that the pretrained representations do not directly expose sufficient 3D structure for driving perception. The head-only setting thus probes how readily the pretrained features support explicit 3D prediction. We then optimize the initialized BEV perception head, vision encoder, and VLM together for this capability. Perception samples use $\mathcal{L}_{\mathrm{perc}}$, while vision-language samples use $\mathcal{L}_{\mathrm{ntp}}$.

Each minibatch contains both sample types. Because they activate different task pathways, we provide dummy inputs to inactive branches to maintain a consistent computation graph across distributed workers. We exclude the corresponding dummy outputs from the loss. The BEV perception head uses a learning rate $20\times$ that of the VLM, allowing the task-specific module to adapt more rapidly during joint training. Driving data provide domain-specific supervision, while general-purpose vision-language data help preserve broad visual understanding and instruction-following capabilities. The resulting VLM representations provide the conditions for the Planning Expert.

\paragraph{Stage 3. Planning Expert Pretraining.}
We keep the vision encoder and VLM fixed and optimize only the Planning Expert. The training mixture contains samples whose prompts include a textual planning reason $\mathbf{r}$ and samples for which $\mathbf{r}=\varnothing$. Both types supervise only the future trajectory through $\mathcal{L}_{\mathrm{plan}}$, with no text-generation objective in this stage. Keeping the conditioning representations fixed separates trajectory learning from changes in the vision-language representations. We refer to the resulting model as \textbf{\ours{}-SFT}.

\paragraph{Stage 4. Reinforcement Learning.}
Stage 3 trains the Planning Expert to reproduce a single recorded future per scene using $\mathcal{L}_{\mathrm{plan}}$. This imitation objective provides stable trajectory supervision but only partially reflects how a plan is evaluated in practice. A recorded trajectory represents only one of several acceptable futures, so other safe behaviors may be penalized, particularly when the four training sources exhibit different ego-motion distributions. Moreover, trajectory regression does not explicitly capture collision avoidance, drivable-area compliance, progress, or agreement with human preference. This stage therefore optimizes the Planning Expert with task-level rewards that measure these properties. We keep the vision encoder and VLM fixed, confining the adaptation to the Planning Expert and preserving the shared representations learned in Stage 2. We refer to the resulting model as \textbf{\ours{}-RL}.

These task-level rewards are nondifferentiable through trajectory generation and therefore require sampled rollouts for optimization. The inference sampler in Sec.~\ref{sec:arch} initializes the trajectory tokens from Gaussian noise and then applies deterministic Euler integration. Once the initial noise is drawn, the remaining integration path is deterministic and defines no transition probabilities that a policy gradient could differentiate. We therefore share the initial trajectory noise within each rollout group and introduce stochastic transitions over a contiguous block of the final integration steps. This converts the deterministic flow into a stochastic policy whose transition likelihood depends on the Planning Expert parameters. Indexing the $K=10$ Euler steps from zero, with $t_k=k/K$ and $\Delta t=1/K$, we introduce stochasticity only over the final three transitions, $\mathcal{W}=\{7,8,9\}$. Under the endpoint parameterization,  perturbations near $t=1$ affect the emitted trajectory more directly, while earlier perturbations are increasingly attenuated by subsequent integration steps. Concentrating exploration near the output therefore yields effective trajectory diversity while limiting deviation from the pretrained flow. Stochastic perturbations can move an intermediate trajectory away from regions favored by the pretrained flow. We therefore construct an approximate restoring score from the Gaussian conditional associated with the interpolation in Eq.~\ref{eq:flowinterp}. Under the Gaussian conditional implied by Eq.~\ref{eq:flowinterp}, we substitute the predicted endpoint $\hat{\boldsymbol{\tau}}_1^{(k)}$ for the unknown clean trajectory and obtain the score correction:
\begin{equation}
s_\theta
\left(
\boldsymbol{\tau}^{(k)}, t_k
\right)
=
\nabla_{\boldsymbol{\tau}^{(k)}}
\log
p_{t_k}\!\left(
\boldsymbol{\tau}^{(k)}
\mid
\hat{\boldsymbol{\tau}}_1^{(k)}
\right)
=
-
\frac{
\boldsymbol{\tau}^{(k)}
-
t_k\hat{\boldsymbol{\tau}}_1^{(k)}
}{
(1-t_k)^2
}.
\label{eq:rl_score}
\end{equation}
This score points toward the conditional center and stabilizes perturbed states. Let $\sigma_k$ denote the standard deviation of one discrete stochastic transition. For a continuous diffusion coefficient $g(t)$, the corresponding discrete standard deviation is $\sigma_k=g(t_k)\sqrt{\Delta t}$. The score drift accumulated over one Euler interval is therefore $\frac{1}{2}g(t_k)^2s_\theta\Delta t=\frac{1}{2}\sigma_k^2s_\theta$. The resulting transition mean is:
\begin{equation}
\boldsymbol{\mu}^{(k)}
=
\boldsymbol{\tau}^{(k)}
+
v_\theta
\left(
\boldsymbol{\tau}^{(k)}, t_k
\right)\Delta t
+
\frac{\sigma_k^2}{2}
s_\theta
\left(
\boldsymbol{\tau}^{(k)}, t_k
\right).
\label{eq:rl_drift}
\end{equation}
We set $\sigma_k=\sigma=0.03$ for $k\in\mathcal{W}$ and $\sigma_k=0$ otherwise. Since $\sigma_k$ denotes the standard deviation of the discrete transition, $\sigma_k^2$ already incorporates the integration interval and requires no additional factor of $\Delta t$ in the score correction. In implementation, $1-t_k$ is lower-bounded by $\epsilon=0.1$, and $\hat{\boldsymbol{\tau}}_1^{(k)}$ is clipped to $[-1,1]$ in normalized coordinates before computing the flow velocity and restoring score.

Independent waypoint noise primarily introduces high-frequency jitter rather than meaningful maneuver diversity, making the resulting samples poorly suited to comparisons of driving quality. We therefore restrict stochastic exploration to a smooth low-frequency temporal subspace. Let $\boldsymbol{\Phi}\in\mathbb{R}^{N\times M}$ contain the first $M=6$ orthonormal cosine modes over the $N=50$ future waypoints, with $\boldsymbol{\Phi}^{\top}\boldsymbol{\Phi}=\mathbf{I}_M$. At each stochastic transition, we sample $\mathbf{Z}_k\in\mathbb{R}^{M\times3}$ with independent standard Gaussian entries and update by:
\begin{equation}
\boldsymbol{\tau}^{(k+1)}
=
\boldsymbol{\mu}^{(k)}
+
\sigma_k\boldsymbol{\Phi}\mathbf{Z}_k.
\label{eq:rldct}
\end{equation}
The low-frequency modes vary smoothly over the prediction horizon, so their combinations produce coherent shifts and bends in the trajectory rather than pointwise oscillations. Since $\boldsymbol{\Phi}$ has orthonormal columns, the perturbation of an individual waypoint has an average standard deviation of $\sigma\sqrt{M/N}\approx0.010$ across waypoints, corresponding to roughly $1.7$\,m longitudinally and $0.26$\,m laterally per stochastic step. The perturbation lies in the $3M$-dimensional subspace of the full $3N$-dimensional trajectory space. We therefore evaluate a Gaussian likelihood surrogate for the injected stochastic action in the corresponding low-dimensional basis coordinates, rather than treating the transition as a full-rank density in trajectory space. Using $\boldsymbol{\Phi}^{\top}\boldsymbol{\Phi}=\mathbf{I}_M$, this likelihood surrogate takes the form:
\begin{equation}
\log
\pi_\theta
\left(
\boldsymbol{\tau}^{(k+1)}
\mid
\boldsymbol{\tau}^{(k)}
\right)
=
-
\frac{1}{2\sigma_k^2}
\left\|
\boldsymbol{\Phi}^{\top}
\left(
\boldsymbol{\tau}^{(k+1)}
-
\boldsymbol{\mu}^{(k)}
\right)
\right\|_F^2
+
\mathrm{const}.
\label{eq:rllogprob}
\end{equation}
The implementation averages the squared residual over the $3M$ mode coefficients instead of summing them, which rescales $\mathcal{L}_{\mathrm{rl}}$ by a constant factor of $1/(3M)$ and is absorbed into the learning rate. Since the score correction in Eq.~\ref{eq:rl_drift} is derived for isotropic diffusion while our perturbation is restricted to a low-dimensional subspace, we interpret it as an approximate restoring correction rather than an exact marginal-preserving transformation.

For each scene, the frozen VLM samples $G=8$ reasoning traces, whose cached keys and values independently condition the $G$ trajectory rollouts from the Planning Expert. The group rollouts share the same initial trajectory noise $\boldsymbol{\tau}^{(0)}$ and sample independent low-frequency perturbations within $\mathcal{W}$, leaving stochastic transitions and sampled reasoning as the sources of within-group diversity. Given rollout rewards $\{R_i\}_{i=1}^{G}$, we compute the group-relative advantage as:
\begin{equation}
A_i
=
\frac{
R_i-\bar{R}
}{
\sigma_R+\epsilon_R
},
\qquad
\bar{R}
=
\frac{1}{G}\sum_{j=1}^{G}R_j,
\label{eq:rl_advantage}
\end{equation}
where $\sigma_R$ is the population standard deviation over the group and $\epsilon_R$=1e-8 is used for numerical stability. This group-relative advantage provides a baseline without a learned value function~\citep{shao2024deepseekmath}. During optimization, the sampled states and advantages are treated as constants, while the transition means are recomputed with the current Planning Expert. Only the stochastic transitions in $\mathcal{W}$ contribute to the objective. Writing $k_w$ for the $w$-th element of $\mathcal{W}$ in ascending order, we optimize the Planning Expert with a discounted policy gradient over the $W=|\mathcal{W}|$ stochastic steps:
\begin{equation}
\mathcal{L}_{\mathrm{rl}}
=
-\frac{1}{GW}
\sum_{i=1}^{G}
\sum_{w=0}^{W-1}
\gamma^{\,W-1-w} A_i
\log \pi_\theta\!\left(\boldsymbol{\tau}_i^{(k_w+1)} \mid \boldsymbol{\tau}_i^{(k_w)}\right),
\label{eq:rlloss}
\end{equation}
where $\gamma=0.6$ assigns greater credit to the steps closest to the output. Each rollout group is sampled and consumed by a single on-policy update. The objective therefore uses the current model log-likelihood directly and requires no off-policy importance correction.

Multi-source reinforcement learning must accommodate each benchmark's distinct evaluation criteria. We use the Predictive Driver Model Score (PDMS) for NAVSIM and the Rater Feedback Score for WOD-E2E, and add a shared displacement term to each source so that a single policy receives a comparable learning signal from all three. Appendix~\ref{app:rl_reward} provides the exact reward definitions. Training uses 15K NAVSIM scenes drawn from \texttt{navtrain} and balanced across navigation commands, 15K PhysicalAI-AV (PAI-AV) scenes, and 479 WOD-E2E scenarios with rater-preference annotations.

\subsection{Data Recipe}
\label{sec:data}

We organize the training data into perception, vision-language, and planning groups and align heterogeneous task definitions within each group before mixing sources.

\subsubsection{Perception Data}
We use nuScenes~\citep{caesar2020nuscenes} and OpenScene~\citep{openscene2023} for single-frame surround-view perception. For nuScenes, we use the semantic occupancy labels provided by nuScenes-OccNet~\citep{tong2023scene}. nuScenes provides six camera views, whereas OpenScene provides eight. We adopt the official nuScenes split, which provides 28K annotated keyframes for training and 6K for validation. For OpenScene, we hold out 16 logs from the trainval pool to balance city and time of day. This split provides 607K training frames and 9K validation frames.

Joint training requires complementary unification at the data and model levels. At the data level, we must reconcile task taxonomies across sources. At the model level, the predicted features must be aligned with dataset-specific spatial grids and coordinate systems. We describe these two procedures below.

\begin{figure*}[t]
\centering
\begin{minipage}[t]{0.43\textwidth}
\vspace{0pt}
\centering
\textbf{(a) Task-specific unification}
\fontsize{7.4}{8.5}\selectfont
\renewcommand{\arraystretch}{1.}
\renewcommand{\tabularxcolumn}[1]{m{#1}}
\begin{tabularx}{\linewidth}{@{}>{\centering\arraybackslash}m{0.25\linewidth}L>{\centering\arraybackslash}m{0.14\linewidth}@{}}
\toprule
\textbf{Task} & \textbf{Cross-dataset processing} & \makecell{\textbf{\# Classes}} \\
\midrule
3D detection
& Merge five vehicle classes, combine bicycle and motorcycle, and retain the OpenScene-only \texttt{czone\_sign} class.
& 7 \\
\midrule
Semantic occupancy
& Apply dataset-specific lookup tables. Complete \texttt{generic\_object} from boxes and \texttt{driveable} from maps.
& 10 \\
\midrule
\makecell{BEV map\\segmentation}
& Online rasterization of both vector maps under a shared schema.
& 6 \\
\bottomrule
\end{tabularx}
\end{minipage}
\hfill
\begin{minipage}[t]{0.56\textwidth}
\vspace{0pt}
\centering
\textbf{(b) Occupancy label processing}
\includegraphics[width=\linewidth]{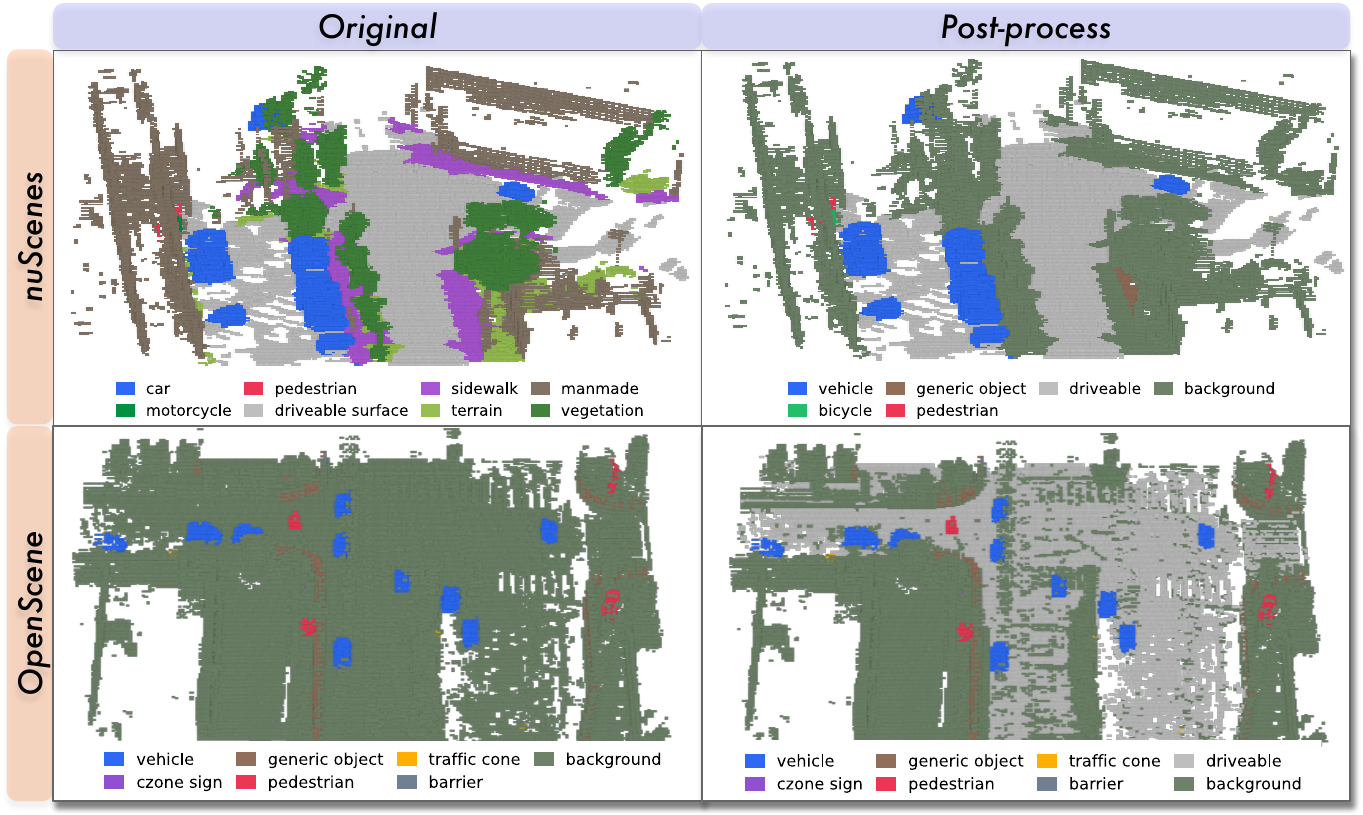}
\end{minipage}
\caption{
    Cross-dataset label unification for perception. (a) Task-specific alignment and label-completion strategies. (b) Original and processed occupancy labels for nuScenes (top) and OpenScene (bottom).
}
\label{fig:perception_data}
\end{figure*}

\paragraph{Label unification.}
We first establish shared task taxonomies across the two sources. Their annotations differ in granularity and class coverage, which prevents direct mixing. We therefore align the taxonomies at the coarsest mutually compatible granularity. Categories annotated by only one source remain source-specific. The loss for each such class is computed only on samples from its source unless reliable auxiliary annotations permit offline completion for the other source. Fig.~\ref{fig:perception_data} summarizes these task-specific procedures and shows representative occupancy labels before and after processing. The corresponding label spaces and processing rules are detailed below.

\begin{itemize}[leftmargin=1.2em]
    \item \textbf{3D detection.} We adopt the annotation granularity of OpenScene. The seven classes are \texttt{vehicle}, \texttt{bicycle}, \texttt{generic\_object}, \texttt{pedestrian}, \texttt{traffic\_cone}, \texttt{barrier}, and \texttt{czone\_sign}. For nuScenes, we merge car, truck, trailer, bus, and construction vehicle into \texttt{vehicle}, and merge bicycle and motorcycle into \texttt{bicycle}. We also map debris, pushable and pullable objects, bicycle racks, animals, and related categories to \texttt{generic\_object}. The \texttt{pedestrian}, \texttt{traffic\_cone}, and \texttt{barrier} classes correspond directly across the two datasets, whereas only OpenScene samples supervise \texttt{czone\_sign}.

    \item \textbf{Semantic occupancy.} Dataset-specific lookup tables map the 17 nuScenes classes and the original OpenScene labels into a shared ten-class space. This space comprises the seven detection classes, \texttt{driveable}, \texttt{background}, and \texttt{empty}. For nuScenes, we map the five vehicle categories to \texttt{vehicle}, bicycle and motorcycle to \texttt{bicycle}, and driveable surface to \texttt{driveable}. Categories without cross-dataset correspondence, including other flat surfaces, sidewalks, terrain, man-made structures, and vegetation, are mapped to \texttt{background}. Free space is mapped to \texttt{empty}. For OpenScene, we map the foreground classes directly. Background surfaces and reserved labels are mapped to \texttt{background}, while unknown and free-space labels are mapped to \texttt{empty}.

    \item \textbf{Offline label completion.} We use reliable auxiliary annotations to complete categories missing from one source, without applying ray masking. The original OpenScene occupancy labels do not distinguish \texttt{driveable}. We rasterize the nuPlan vector map and relabel a voxel as \texttt{driveable} only if it is a ground voxel inside a driveable region and was originally labeled \texttt{background}. nuScenes does not provide occupancy annotations for \texttt{generic\_object}. We generate pseudo-labels from the 3D boxes of bicycle racks, debris, and pushable and pullable objects. Within these boxes, we change only voxels that already carry semantic labels. After remapping and completion, we recompute class frequencies in the unified label space for the class-balanced focal loss.

    \item \textbf{BEV map segmentation.} We rasterize both vector maps online under a shared six-class schema instead of remapping existing raster labels. The schema contains driveable surface, road line, road edge, crosswalk, walkway, and background.
\end{itemize}

Label unification does not remove noise from the source annotations. nuScenes provides manually annotated semantics, whereas OpenScene reconstructs occupancy targets from aggregated LiDAR sweeps using an automated pipeline without per-point semantics. Sensor and registration errors can therefore introduce artifacts, including floating voxels detached from physical surfaces. Offline completion adds missing semantic labels but retains these artifacts.

\paragraph{Spatial unification.}
Model-level unification poses a separate spatial dilemma. Both sources store occupancy as $200\times200\times16$ voxel grids, yet corresponding voxel indices represent different physical locations. nuScenes-OccNet spans $\pm40$\,m horizontally and $z\in[-1.0,5.4]$\,m with 0.4\,m voxels, while OpenScene spans $\pm50$\,m and $z\in[-4.0,4.0]$\,m with 0.5\,m voxels. Their coordinate transformations also differ. The nuScenes LiDAR has a non-identity transform to the rear-axle ego frame, including a vertical offset of $\sim$1.84\,m, while OpenScene uses an identity LiDAR-to-ego transform. Directly sharing voxel indices would misalign the two sources, while resampling categorical labels would distort the supervision. We therefore preserve each source's native occupancy grid and perform spatial alignment on the predicted features.

The depth-based view transform first lifts image features into a 3D volume defined over the detection range. Before occupancy decoding, a single differentiable trilinear sampling operation maps this volume onto the dataset-specific occupancy grid. For nuScenes, each target voxel center is defined in the ego frame and transformed back into the LiDAR frame for sampling. The same operation restricts the output to the occupancy range, while an expanded vertical source range of $[-5.0,5.4]$\,m covers the LiDAR mounting offset. For OpenScene, the identity transform requires no frame conversion. The head selects the corresponding occupancy range and voxel size for each source during the forward pass. The same occupancy head can therefore predict on both native grids without dataset-specific branches. Both sources supervise this head under a consistent ego-frame convention while retaining their native spatial resolution. BEV map supervision is unified separately. For both sources, we rasterize the vector maps online over an ego-centered local patch with $x\in[-30,30]$\,m and $y\in[-15,15]$\,m at 0.15\,m resolution, producing a $400\times200$ target rather than rasterizing the full city map.

\subsubsection{Vision-Language Data}
The vision-language data combine general-purpose and driving examples. The driving component covers scene understanding, spatial grounding, cross-view reasoning, and planning reasoning.

\paragraph{Data Sources.}
The driving component combines open-source datasets with self-constructed examples that target underrepresented tasks. We detail the composition and preprocessing of these two source groups below. Fig.~\ref{fig:vl_data} shows the filtered public driving data, their scene diversity, and the Stage 2 mixture.

\begin{itemize}[leftmargin=1.2em]
    \item \textbf{Open-Source Driving Data.} We aggregate 24 publicly available driving vision-language datasets, including CODA-LM~\citep{chen2025automated}, DRAMA~\citep{malla2023drama}, DriveAction~\citep{hao2025driveaction}, DriveGPT4~\citep{xu2024drivegpt4}, DriveLM~\citep{sima2024drivelm}, DrivingVQA~\citep{corbiere2026drivingvqa}, Impromptu VLA~\citep{chiimpromptu}, LingoQA~\citep{marcu2024lingoqa}, MapLM~\citep{cao2024maplm}, MM-AU~\citep{fang2024abductive}, NAVSIM-ReCogDrive~\citep{li2025recogdrive}, NuInstruct~\citep{ding2024holistic}, NuPlanQA~\citep{park2025nuplanqa}, nuScenes-MQA~\citep{inoue2024nuscenes}, nuScenes-QA~\citep{qian2024nuscenes}, the OOD reasoning-label subset of PhysicalAI-AV~\citep{nvidia2025physicalai}, OmniDrive~\citep{wang2025omnidrive}, ROADWork~\citep{ghosh2025roadwork}, Senna~\citep{jiang2024senna}, STSBench~\citep{fruhwirth2025stsbench}, SURDS~\citep{guo2025surds}, SUTD-TrafficQA~\citep{xu2021sutd}, Talk2Car~\citep{deruyttere2019talk2car}, and WaymoQA~\citep{yu2025waymoqa}. We use training splits for these datasets to avoid data leakage.

    The released datasets differ in conversational format and annotation reliability, and many targets originate from templates or automated pipelines. Before mixing the datasets, we use Qwen3.5-Plus~\citep{qwen35blog} to rewrite each prompt and response into a common conversational schema. The source annotation remains the reference for subsequent consistency filtering. We convert most multiple-choice questions to open-ended QA but retain a subset in multiple-choice form to preserve this instruction type. We normalize bounding boxes to $[0,1000)$ image coordinates, insert view and frame tags following Sec.~\ref{sec:arch}, and revise textual view references to match the inserted tags.

    Rewriting standardizes the format but does not validate the source annotation. We therefore apply a separate consistency filter. Qwen3.5-Flash evaluates whether each rewritten response is semantically consistent with its source annotation, and we retain only samples classified as consistent. As shown in Fig.~\ref{fig:vl_data}(a), this procedure reduces the data from 5.53M to 3.09M samples, corresponding to a retention rate of 55.9\%. The retained set contains 61.6\% multi-view, 20.1\% single-view, 10.1\% single-view temporal, 4.4\% multi-view temporal, and 3.7\% video samples. It covers scene and region captioning, open-ended and multiple-choice QA, 2D grounding, spatial reasoning, and planning reasoning. The representative samples in Fig.~\ref{fig:vl_data}(c) further illustrate the diversity of road environments, illumination, and weather conditions.

    \begin{figure*}[t]
    \centering
    \includegraphics[width=\textwidth]{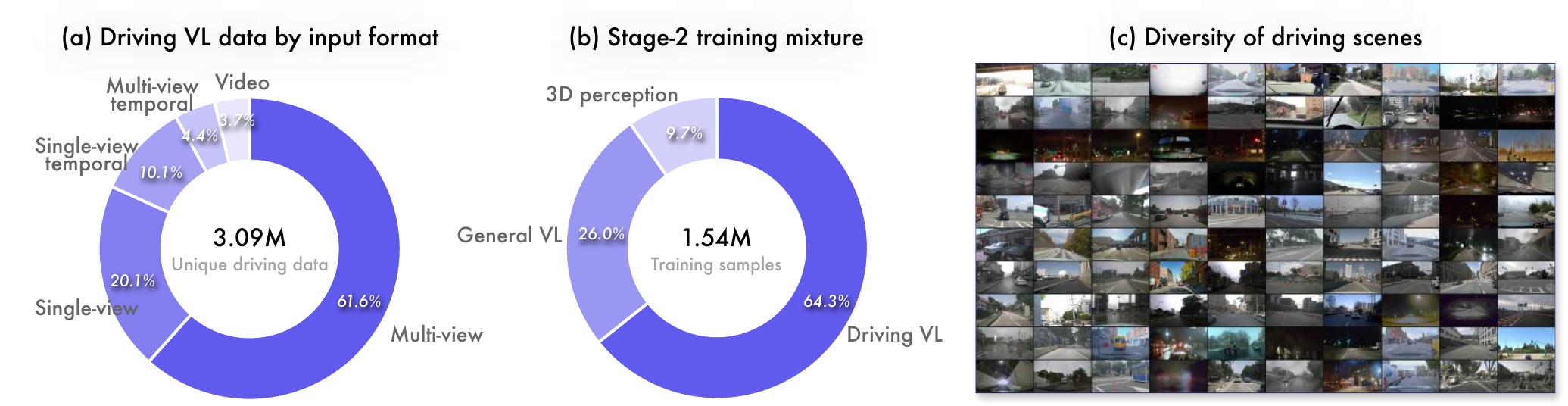}
    \caption{
        Vision-language data and the Stage 2 training mixture. (a) Input-format distribution of the 3.09M filtered public driving samples before Stage 2 subsampling. (b) Composition of the 1.54M Stage 2 training set before repetition. (c) Representative scenes spanning diverse road environments, illumination, and weather conditions.
    }
    \label{fig:vl_data}
    \end{figure*}

    \item \textbf{Self-Constructed Driving Data.} We construct three additional components to provide supervision that is underrepresented in the public datasets.

    \textbf{1)} Planning reasoning. We construct planning-reasoning data using a Chain-of-Causation (CoC) formulation inspired by Alpamayo-R1~\citep{wang2025alpamayo}. A CoC trace identifies the scene elements that motivate a driving decision and explains their causal roles. We build the data from three publicly available driving datasets with future ego trajectories, namely NAVSIM~\citep{dauner2024navsim}, Waymo~\citep{sun2020scalability}, and PAI-AV~\citep{nvidia2025physicalai}. From each future ego trajectory, a rule-based classifier derives longitudinal and lateral maneuver components that jointly form a motion prior. Qwen3.7-Plus~\citep{qwen37plus} then generates a planning-reasoning trace conditioned on the multi-view images, historical ego trajectory, motion prior, and navigation instruction.

    Each generated trace undergoes a multistage audit. Rather than requesting scalar quality scores, judge models answer classification questions, and their decisions are aggregated programmatically. The audit checks the predicted maneuver against the ground-truth trajectory, classifies the causal role of each cited factor, and rejects traces that reveal future information. Qwen3.5-Flash further assigns a rarity score to each scene, and rare scenes receive higher sampling priority. For each accepted trace, we construct two response formats. The first contains only the planning-reasoning trace. The second contains the same trace followed by the future ego trajectory serialized in JSON format.

    \textbf{2)} Camera ordering for cross-view spatial understanding. We shuffle the surround-view images and remove all view tags. The model identifies the front view from visual cues and recovers the clockwise order of the complete camera set.

    \textbf{3)} In-house perception QA. We construct 30K examples from road scenes collected in China for traffic-light grounding and 3D object detection. In the 3D detection examples, the camera pose is provided in text, and the response contains 3D boxes in the global coordinate frame.
\end{itemize}

The public sources provide broad coverage of scenes and question types, while the self-constructed data provide targeted supervision for causal reasoning and perception grounding.

\paragraph{Stage 2 Data Mixture.}
Stage 2 mixes the perception and vision-language data described above. Because the filtered public driving data exceed the budget for this stage, we sample $\sim$20\% from each public source, stratified by task and question type. We combine this subset with the self-constructed driving data and general-purpose vision-language data. Before repetition, the resulting set contains 1.54M examples. As shown in Fig.~\ref{fig:vl_data}(b), 9.7\% provide 3D perception supervision, 26.0\% provide general-purpose vision-language supervision, and 64.3\% provide driving vision-language supervision.

We use group-specific repetition factors, assigning a larger factor to perception examples to increase updates to the BEV perception head and repeating vision-language examples for two to three epochs. After repetition, the effective mixture contains 12.7\% perception, 31.0\% general-purpose vision-language, and 56.3\% driving vision-language supervision.

\subsubsection{Planning Data}
For Stage 3, we use NAVSIM~\citep{dauner2024navsim}, OpenScene~\citep{openscene2023}, WOD-E2E~\citep{xu2026wod}, and PAI-AV~\citep{nvidia2025physicalai}. NAVSIM and OpenScene are both derived from the nuPlan dataset~\citep{karnchanachari2024towards}, but remain separate sources because their ego-motion distributions differ. The resulting training set contains $\sim$2.83M samples. NAVSIM and OpenScene jointly contribute 890K samples from 2.5K clips. WOD-E2E contributes 557K samples from 2K clips, while PAI-AV contributes 1.38M samples from 156K clips. Of the training samples, 685K (24.2\%) include an accepted planning-reasoning trace as a condition. These samples comprise 78K from NAVSIM, 142K from WOD-E2E, and 465K from PAI-AV. The remaining 75.8\% omit this condition and use $\mathbf{r}=\varnothing$.

During preprocessing, we express every future trajectory in the current ego frame and convert it to the 50-waypoint representation in Eq.~\ref{eq:traj}. For NAVSIM and OpenScene, we read positions and headings at 10\,Hz directly from the nuPlan database. For PAI-AV, we recover future ego motion from the per-clip egomotion annotations and resample it on the same temporal grid. Because this source is far larger than the others, we retain only a small number of evenly spaced frames from each clip.

WOD-E2E provides future ego positions at 4\,Hz. We prepend the current position, fit a time-parameterized natural cubic spline to the position sequence, and evaluate it on the 10\,Hz grid. Its first and second derivatives provide the future velocities and accelerations. For historical acceleration, we apply a factor-of-four scale correction to the raw \texttt{accel\_x} and \texttt{accel\_y} metadata and linearly resample the corrected values at 10\,Hz. Let $\mathbf{v}=(v_x,v_y)$ and $\mathbf{a}=(a_x,a_y)$ denote the spline-derived velocity and acceleration. For $\|\mathbf{v}\|\geq0.3\,\mathrm{m/s}$, the induced heading rate is $\dot{\theta}=(v_xa_y-v_ya_x)/\|\mathbf{v}\|^2$. We set $\dot{\theta}=0$ below this speed, integrate it on the 10\,Hz grid, and set the current heading to zero. The resulting future headings provide the $\theta_k$ values in Eq.~\ref{eq:traj}. We retain samples only when the historical and future acceleration magnitudes do not exceed standard gravity ($9.8\,\mathrm{m/s^2}$). For the unwrapped future heading sequence, we apply complementary derivative-based and adjacent-step checks to capture both rate consistency and abrupt local changes. Both checks use a threshold of $1.2\,\mathrm{rad/s}$.

Each planning example contains front, front-left, and front-right images at four timesteps. These comprise the current and three historical timesteps sampled at 0.5\,s. The images form the vehicle sensor inputs $\mathbf{s}$. The historical ego trajectory $\boldsymbol{\tau}_{\mathrm{hist}}$, current ego state $\mathbf{e}$, and navigation instruction $\mathbf{n}$ provide the remaining conditions in Eq.~\ref{eq:traj}. Accepted planning-reasoning traces provide $\mathbf{r}$ for a subset of samples, while $\mathbf{r}=\varnothing$ for the remainder. For these samples, $\ell$ is the view-major layout defined in Sec.~\ref{sec:arch}. We resize historical images to 320p and current images to 720p. This allocation retains more spatial detail in the current observation while representing the motion history with fewer visual tokens.

\section{Experiments}
\label{sec:exp}

We evaluate \ours{} across 3D perception, driving, general vision-language understanding, and motion planning. The perception and vision-language evaluations use \ours{}-SFT, while we evaluate motion planning with both \ours{}-SFT and \ours{}-RL.

\subsection{3D Perception}
\label{sec:exp_perception}

\paragraph{Metrics.}
We evaluate on the official nuScenes validation set and the 16-log OpenScene validation split defined in Sec.~\ref{sec:data}. For 3D detection, we report mean average precision (mAP) and a modified nuScenes detection score (NDS) over the unified seven-class label space. We match a prediction to a ground-truth box when their BEV center distance is below a specified threshold. We compute AP as the normalized area under the precision-recall curve after discarding operating points with precision or recall below 10\%. The mAP averages AP over the center-distance thresholds $\{0.5,1,2,4\}$\,m and all classes. For NDS, we compute the true-positive error metrics at the 2\,m threshold following the official class-specific definitions. Because the unified annotations contain no attribute labels, our modified NDS sets the mean attribute error (mAAE) to zero for all methods.

For semantic occupancy, we report the mean IoU over the unified semantic classes excluding \texttt{empty} (Occ mIoU), and RayIoU~\citep{liu2024fully}. Notably, OpenScene stores occupancy in the rear-axle ego frame with an identity LiDAR-to-ego transform. Rays cast from the rear axle would therefore originate at road level and immediately terminate on road-surface voxels. We instead raise the OpenScene ray origin by 1.84\,m to match the nuScenes LiDAR mounting height. For BEV map segmentation, we report the mean IoU over the foreground map classes.

\paragraph{Comparison Methods.}
To enable fair comparisons, we reproduce the single-frame variants of BEVFormerV2~\citep{yang2023bevformer}, PETR~\citep{liu2022petr}, and PETRv2~\citep{liu2023petrv2} on the remapped nuScenes annotations. Our BEVFormerV2 implementation excludes the Group-DETR decoder and its associated 2D losses. We further construct BEVFormerV2$^{*}$ as a unified multi-task variant that jointly performs 3D detection, BEV map segmentation, and semantic occupancy prediction, covering the same three perception tasks as our BEV perception head. We evaluate each architecture with ResNet-50~\citep{he2016deep} and SigLIP-Qwen. SigLIP-Qwen denotes the SigLIP-style~\citep{zhai2023sigmoid} vision encoder initialized with the same pretrained Qwen3.5-4B weights used by \ours{}. All comparison methods follow a 24-epoch schedule. Both the comparison methods and our BEV perception head use an input resolution of $896\times512$. In contrast, our BEV perception head uses no rig-specific camera embeddings, so a single model trains and evaluates across the six-camera nuScenes rig and the eight-camera OpenScene rig.

\begin{table*}[t]
\centering
\caption{
    Unified 3D perception on the remapped nuScenes and OpenScene validation splits. Comparison methods are trained only on remapped nuScenes under a common schedule and input resolution. Dashes indicate tasks that a method does not perform. The comparison methods learn camera embeddings tied to the six-camera nuScenes rig and therefore cannot be evaluated on the eight-camera OpenScene rig. Gray values denote cross-dataset evaluation of the nuScenes-only head on OpenScene. NDS denotes the modified score with mAAE set to zero because the unified labels contain no attributes. $^{*}$ denotes the unified multi-task BEVFormerV2 variant with map segmentation and occupancy prediction.
}
\label{tab:perception}
\footnotesize
\setlength{\tabcolsep}{2.8pt}
\begin{tabular}{ll ccccc ccccc}
\toprule
\multirow{3.6}{*}{Method} & \multirow{3.6}{*}{Visual Encoder} & \multicolumn{5}{c}{nuScenes} & \multicolumn{5}{c}{OpenScene} \\
\cmidrule(lr){3-7} \cmidrule(lr){8-12}
& & \multicolumn{2}{c}{3D Det} & Map & \multicolumn{2}{c}{Occ} & \multicolumn{2}{c}{3D Det} & Map & \multicolumn{2}{c}{Occ} \\
\cmidrule(lr){3-4} \cmidrule(lr){5-5} \cmidrule(lr){6-7} \cmidrule(lr){8-9} \cmidrule(lr){10-10} \cmidrule(lr){11-12}
& & mAP & NDS & mIoU & mIoU & RayIoU & mAP & NDS & mIoU & mIoU & RayIoU \\
\midrule
BEVFormerV2 & ResNet-50 & 33.04 & 33.02 & -- & -- & -- & \multicolumn{5}{c}{--} \\
PETR & ResNet-50-DCN & 29.77 & 26.65 & -- & -- & -- & \multicolumn{5}{c}{--} \\
PETRv2 & ResNet-50-DCN & 25.98 & 23.90 & 52.52 & -- & -- & \multicolumn{5}{c}{--} \\
BEVFormerV2$^{*}$ & ResNet-50 & 35.34 & 30.76 & 48.49 & 23.39 & 40.69 & \multicolumn{5}{c}{--} \\
\midrule
BEVFormerV2 & SigLIP-Qwen & 40.78 & 39.78 & -- & -- & -- & \multicolumn{5}{c}{--} \\
PETR & SigLIP-Qwen & 37.61 & 34.37 & -- & -- & -- & \multicolumn{5}{c}{--} \\
PETRv2 & SigLIP-Qwen & 36.10 & 33.28 & 57.62 & -- & -- & \multicolumn{5}{c}{--} \\
BEVFormerV2$^{*}$ & SigLIP-Qwen & 41.94 & 36.46 & 47.76 & \textbf{25.72} & \textbf{43.89} & \multicolumn{5}{c}{--} \\
\midrule
Head-only (nuScenes) & SigLIP-Qwen & 35.60 & 34.13 & 55.55 & 20.21 & 36.98 & \textcolor{dt}{16.14} & \textcolor{dt}{16.50} & \textcolor{dt}{40.45} & \textcolor{dt}{11.50} & \textcolor{dt}{17.43} \\
Head-only (joint) & SigLIP-Qwen & 33.49 & 33.37 & 51.15 & 14.83 & 29.54 & 40.57 & 41.86 & 66.34 & \textbf{20.13} & \textbf{25.36} \\
\textbf{\ours{}-SFT} & SigLIP-Qwen & \textbf{43.95} & \textbf{42.83} & \textbf{60.99} & 19.82 & 37.02 & \textbf{43.45} & \textbf{44.16} & \textbf{71.27} & 19.84 & 25.17 \\
\bottomrule
\end{tabular}
\end{table*}

\paragraph{Quantitative Results.}
As shown in Tab.~\ref{tab:perception}, on nuScenes, \ours{}-SFT establishes the best detection and map-segmentation results, exceeding BEVFormerV2$^{*}$ by 2.01 mAP, BEVFormerV2 by 3.05 NDS, and PETRv2 by 3.37 map mIoU. On OpenScene, it also improves over Head-only (joint) in both detection metrics and map mIoU. Replacing ResNet-50 with the pretrained ViT consistently benefits the dedicated detectors, confirming that vision-language pretraining provides a strong visual initialization. However, a converged head trained on the same SigLIP-Qwen features still trails BEVFormerV2$^{*}$ by 6.34 mAP and 6.91 RayIoU. This contrast indicates that vision-language-pretrained features support visual-text alignment but do not directly expose the 3D structure required for driving perception.

Cross-dataset transfer presents a separate challenge. Despite training on the higher-quality nuScenes annotations, the nuScenes-only head reaches only 16.50 NDS on OpenScene, less than half of its 34.13 nuScenes score. Mixed-source training raises OpenScene NDS to 41.86, but the residual differences between label semantics and annotation pipelines introduce negative transfer on nuScenes. The effect is most pronounced for occupancy, where mIoU decreases by 26.6\%. The same source mismatch persists after joint adaptation: Stage 2 substantially recovers nuScenes occupancy but changes both OpenScene occupancy metrics by less than 0.3 points, despite clear improvements in detection and map segmentation. Unlike boxes and rasterized map layers, the machine-generated OpenScene voxel labels retain source-specific semantic and construction artifacts, which limit the benefit of joint adaptation. This observation explains why \ours{}-SFT leads in detection and map segmentation but not occupancy. Label mapping and offline completion therefore make joint training feasible without fully resolving source ambiguity at the voxel level.

The gains from Stage 2 cannot be explained by continued head optimization alone, since the head-only model had already converged. Once the perception objectives are allowed to update the pretrained ViT encoder and VLM, nuScenes mAP and map mIoU increase by 10.46 and 9.84 points, while OpenScene map mIoU gains a further 4.93 points. Joint adaptation is therefore important for realizing the 3D perception capability of the external head. Together with the ablation in Tab.~\ref{tab:ablation_vl}, these results show that the head provides a practical 3D probe and that targeted adaptation produces explicit, inspectable scene predictions while preserving highly competitive vision-language performance. This capability is added without changing the VLM architecture.

\begin{figure*}[t]
\centering
\includegraphics[width=\textwidth]{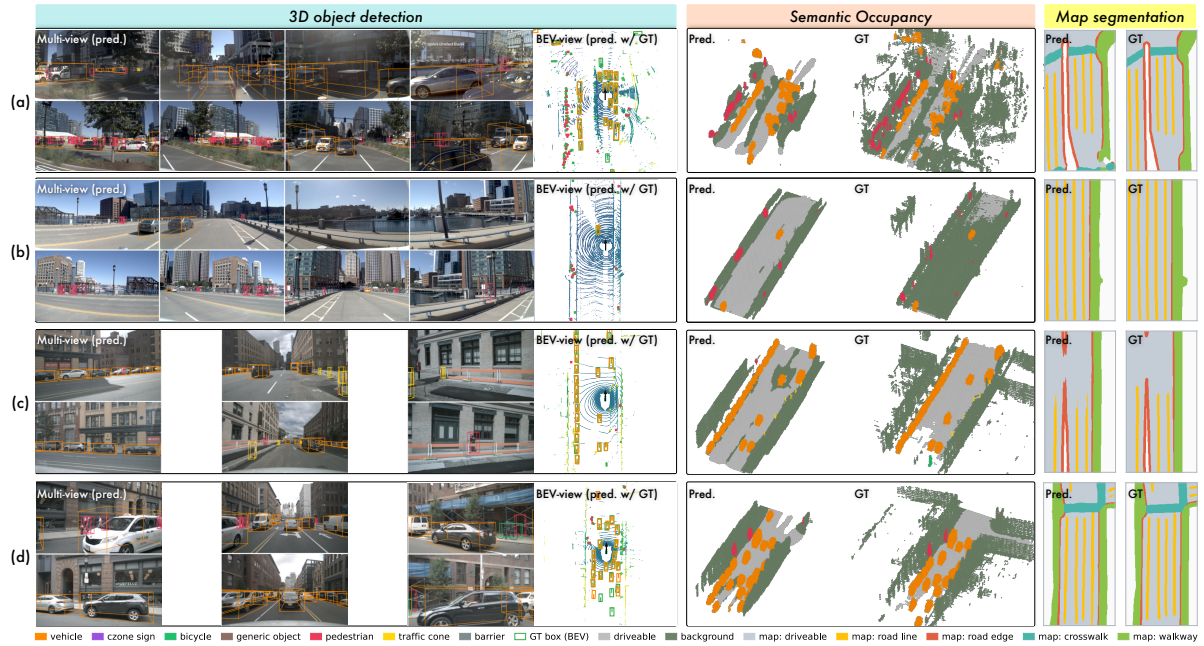}
\caption{
    Qualitative results of \ours{}-SFT on the OpenScene (a, b) and nuScenes (c, d) validation splits. Each row shows 3D detection, semantic occupancy, and BEV map segmentation results.
}
\label{fig:perception_vis}
\end{figure*}

\paragraph{Qualitative Results.}
Fig.~\ref{fig:perception_vis} shows that \ours{}-SFT produces competitive 3D perception results within the camera-visible regions under both the six-view and the eight-view rig. Notably, in Fig.~\ref{fig:perception_vis}(b), our preprocessing does not fully remove floating voxels above the road surface from the machine-generated OpenScene label, while the prediction suppresses these artifacts and correctly recovers the road-surface semantics. This example suggests that supervision across scenes and data sources can reduce sensitivity to residual artifacts in processed labels.

\subsection{Driving and General Vision-Language Understanding}
\label{sec:exp_vqa}

\subsubsection{Driving Visual Question Answering}

\paragraph{Benchmark Selection.}
Existing driving VQA benchmarks present three issues that obscure a model's true driving competence. \textbf{1)} Annotation quality is uneven, and many reference answers cannot be inferred from the visual input alone. \textbf{2)} Text-similarity scores reward surface agreement with the reference wording and therefore measure fit to the training distribution rather than the semantic quality of a response. \textbf{3)} The queried content is often either too fine-grained, such as the exact count of surrounding pedestrians, or too generic, such as an open description of the scene, so neither probes the understanding of traffic participants, driving behavior, scene risk, and road topology.

We therefore select benchmarks with reliable annotations and evaluation based on an LLM judge or a multiple-choice protocol. The five public benchmarks we adopt together emphasize spatial judgment, fine-grained object grounding, driving decisions, and traffic-risk assessment. LingoQA~\citep{marcu2024lingoqa} evaluates free-form driving QA with a judge model.\footnote{We use Qwen-Plus as the judge instead of the official LingoJudge, which we found to score leniently and inconsistently across scenarios. A stronger judge yields a stricter and more reliable assessment. Under the official LingoJudge protocol, \ours{}-SFT obtains a LingoScore of 79.4.} Ego3D-Bench~\citep{gholami2026spatial} evaluates categorical spatial reasoning with multiple-choice accuracy and absolute distance estimation with RMSE. Its distance queries include ego-to-object and cross-view object-to-object relations. VLADBench~\citep{li2025fine} organizes driving competence into a hierarchy of fine-grained capabilities under a multiple-choice protocol. SURDS~\citep{guo2025surds} targets spatial understanding and reasoning for driving. WaymoQA~\citep{yu2025waymoqa} focuses on safety-critical multi-view situations, for which we report the overall score and the safety-specific score for the image split.

In addition to public benchmarks, we introduce PAI-AV-CoC, a CoC benchmark derived from the PAI-AV validation split. Qwen3.5-Plus acts as a judge, extracting key objects and driving decisions from predicted CoC traces. We measure key-object accuracy, decision accuracy, and overall accuracy (requiring both to be correct) to assess the causal content of reasoning traces, rather than just their phrasing. We also evaluate on an in-house Chinese urban driving decision benchmark with expert annotations to assess decision-making ability.

\paragraph{Comparison Methods.}
We compare against recent vision-language models that target physical AI and autonomous driving. The Cosmos family progresses through three generations. Cosmos-Reason1~\citep{azzolini2025cosmos} post-trains Qwen2.5-VL on physical common sense and embodied reasoning with supervised fine-tuning and reinforcement learning. Cosmos-Reason2~\citep{nvidia2025cosmos} extends this recipe to Qwen3-VL at 2B, 8B, and 32B parameters, and Cosmos3-nano~\citep{agarwal2026cosmos} is the most recent omnimodal model. MiMo-Embodied~\citep{hao2025mimo} is a 7B cross-embodied model that builds on MiMo-VL and is pretrained on large-scale autonomous driving and embodied data. UniDriveVLA~\citep{li2026unidrivevla} decouples driving perception, understanding, and planning into separate experts trained progressively on driving data, and we evaluate its checkpoint after the VQA stage and before planning training. Alpamayo-1.5~\citep{wang2025alpamayo} is a 10B reasoning driving model built on Cosmos-Reason2 and trained on chain-of-causation reasoning traces together with large-scale open-source driving data. Beyond these domain-specific models, we further include InternVL3.5-8B-Instruct~\citep{wang2025internvl3}, LLaVA-OneVision-2-8B-Instruct (LLaVA-OV2-8B)~\citep{an2026llava}, and Gemma4-12B~\citep{team2026gemma} as representative general-purpose VLMs. We re-evaluate all comparison methods under a common protocol. Decoding uses near-deterministic sampling with top-$k{=}1$, top-$p{=}0.001$, and temperature $0.01$, without repetition or presence penalties, which reduces generation variance. The same judge scores every method on each benchmark.

\begin{table*}[t]
\centering
\caption{
    Driving VQA results. Higher scores are better, except for Ego3D RMSE. IH denotes the in-house driving-decision benchmark. The first Avg.\ averages the six higher-is-better metrics in the left group and excludes the Ego3D RMSE. The second averages the three PAI-AV-CoC metrics and IH. ``--'' indicates an invalid or unparsable response and is counted as zero in the relevant average. \textbf{Bold} and \underline{underlined} values denote the best and second-best, respectively.
}
\label{tab:driving_vqa}
\footnotesize
\setlength{\tabcolsep}{2.1pt}
\begin{tabular}{l cccccccc ccccc}
\toprule
\multirow{3.6}{*}{Method} & \multicolumn{8}{c}{Driving QA \& Spatial Understanding} & \multicolumn{5}{c}{Causal Reasoning} \\
\cmidrule(lr){2-9} \cmidrule(lr){10-14}
& \multirow{2.5}{*}{LingoQA} & \multicolumn{2}{c}{Ego3D} & \multirow{2.5}{*}{VLAD} & \multirow{2.5}{*}{SURDS} & \multicolumn{2}{c}{WaymoQA} & \multirow{2.5}{*}{Avg.} & \multicolumn{3}{c}{PAI-AV-CoC} & \multirow{2.5}{*}{IH} & \multirow{2.5}{*}{Avg.} \\
\cmidrule(lr){3-4} \cmidrule(lr){7-8} \cmidrule(lr){10-12}
& & Acc. & RMSE$\downarrow$ & & & Safety & All & & Key & Plan & All & & \\
\midrule
InternVL3.5-8B-Inst. & 46.40 & 47.38 & 23.01 & 54.47 & 32.80 & 54.47 & 58.09 & \cellcolor{tabhl}48.94 & -- & -- & -- & 47.50 & \cellcolor{tabhl}11.88 \\
LLaVA-OV2-8B & 41.20 & 42.07 & 24.97 & 58.71 & 38.60 & 49.65 & 55.23 & \cellcolor{tabhl}47.58 & 0.86 & 12.32 & 0.57 & 54.00 & \cellcolor{tabhl}16.94 \\
Gemma4-12B & 50.00 & 56.31 & 27.30 & 57.90 & 52.25 & 63.59 & 68.82 & \cellcolor{tabhl}58.15 & 12.89 & 10.32 & 4.01 & \underline{61.00} & \cellcolor{tabhl}\underline{22.05} \\
Qwen3.5-4B & 70.40 & \textbf{62.82} & 13.17 & \underline{65.38} & \underline{52.95} & 62.46 & 67.10 & \cellcolor{tabhl}\underline{63.52} & 8.88 & 9.17 & 2.58 & 59.00 & \cellcolor{tabhl}19.91 \\
\midrule
Cosmos-Reason1-7B & 45.20 & 44.17 & 26.71 & 33.64 & 8.49 & 39.53 & 43.90 & \cellcolor{tabhl}35.82 & 14.61 & 11.75 & 3.15 & 30.50 & \cellcolor{tabhl}15.00 \\
Cosmos-Reason2-2B & 38.60 & 44.42 & 12.03 & 53.90 & 29.46 & 63.71 & 59.65 & \cellcolor{tabhl}48.29 & 9.17 & 5.44 & 2.01 & 42.50 & \cellcolor{tabhl}14.78 \\
Cosmos-Reason2-8B & 59.60 & 48.35 & 12.62 & 56.37 & 19.54 & 57.68 & 57.93 & \cellcolor{tabhl}49.91 & 7.74 & 5.44 & 1.72 & 56.00 & \cellcolor{tabhl}17.73 \\
Cosmos-Reason2-32B & 58.80 & 47.31 & 20.32 & 57.13 & 19.52 & 48.56 & 48.40 & \cellcolor{tabhl}46.62 & \underline{18.34} & \underline{15.47} & \underline{5.73} & 29.50 & \cellcolor{tabhl}17.26 \\
Cosmos3-nano & 65.00 & 44.02 & 22.41 & 57.73 & 39.72 & 56.93 & 58.36 & \cellcolor{tabhl}53.63 & 12.89 & 10.60 & 4.01 & 2.00 & \cellcolor{tabhl}7.38 \\
MiMo-Embodied-7B & \underline{72.00} & 60.41 & 9.85 & 50.33 & 43.06 & \underline{66.54} & \underline{69.56} & \cellcolor{tabhl}60.32 & -- & -- & -- & \underline{61.00} & \cellcolor{tabhl}15.25 \\
UniDriveVLA-8B & 62.00 & 44.11 & \underline{8.45} & 53.09 & 20.06 & 49.19 & 49.28 & \cellcolor{tabhl}46.29 & 0.86 & 12.89 & 0.86 & 48.50 & \cellcolor{tabhl}15.78 \\
Alpamayo-1.5-10B & 64.00 & 36.79 & 25.31 & 9.13 & 3.10 & 42.61 & 44.37 & \cellcolor{tabhl}33.33 & 9.46 & 8.31 & 3.44 & 3.00 & \cellcolor{tabhl}6.05 \\
\midrule
\textbf{\ours{}-SFT} & \textbf{77.80} & \underline{60.98} & \textbf{7.78} & \textbf{66.52} & \textbf{66.13} & \textbf{70.70} & \textbf{74.47} & \cellcolor{tabhl}\textbf{69.43} & \textbf{65.33} & \textbf{55.59} & \textbf{41.26} & \textbf{71.00} & \cellcolor{tabhl}\textbf{58.30} \\
\bottomrule
\end{tabular}
\end{table*}

\begin{figure*}[t]
\centering
\includegraphics[width=\textwidth]{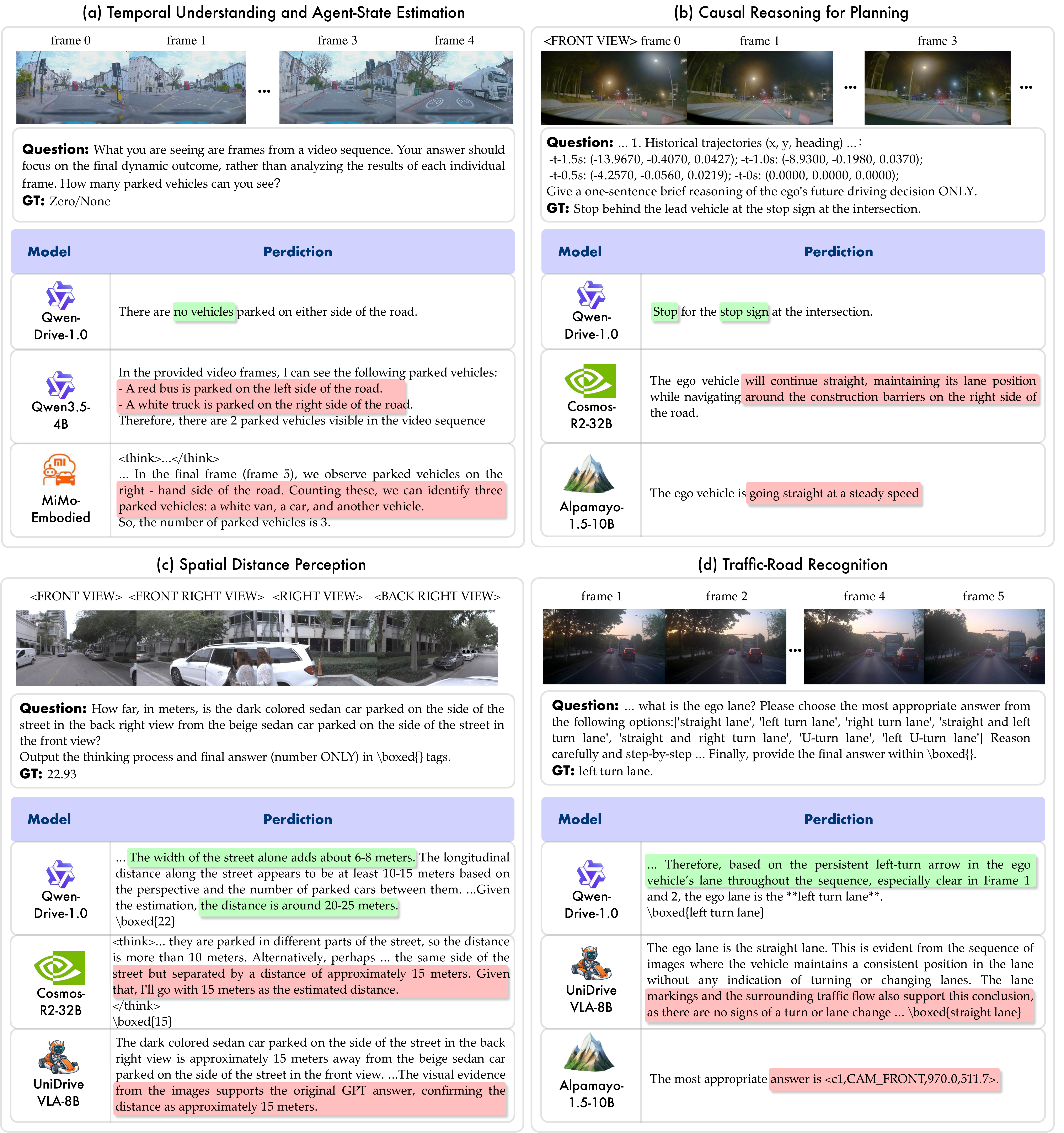}
\caption{
    Qualitative comparison of driving VQA capabilities. Green and red highlight correct and incorrect content, respectively. Questions and responses are abridged for space, with complete examples provided in Appendix~\ref{app:drivevqa_cases}.
}
\label{fig:drivevqa_vis}
\end{figure*}

\paragraph{Results.}
As shown in Tab.~\ref{tab:driving_vqa}, Qwen3.5-4B is the strongest general-purpose reference despite being the smallest, and its driving QA average of 63.52 surpasses the larger Gemma4-12B at 58.15 as well as every specialized comparison method. \ours{}-SFT improves this average by 5.91 points to 69.43, the highest among all methods. This improvement is not concentrated in a single skill but distributed across complementary capabilities. LingoQA rises by 7.40 points and SURDS by 13.18 points, a 24.9\% relative gain, while the Ego3D distance-estimation RMSE drops by 40.9\% to 7.78, which is 7.9\% below the 8.45 of UniDriveVLA-8B. We attribute this reduction not to geometrically precise representations learned by the VLM, but to a richer physical understanding of driving scenes together with the model's retained quantitative reasoning. Semantic cues of stable real-world scale, such as the regular spacing of parked cars along a street or of dashed lane markings and streetlights, provide implicit references for metric distance estimation, as illustrated in Fig.~\ref{fig:drivevqa_vis}(c).

The most pronounced gains appear in causal reasoning, where \ours{}-SFT attains an average of 58.30 versus 22.05 for the second-best Gemma4-12B. Relative to Qwen3.5-4B, \ours{}-SFT improves key-object, decision, and overall accuracy by 56.45, 46.42, and 38.68 points, respectively. Its overall accuracy of 41.26 is more than seven times that of the much larger Cosmos-Reason2-32 B (5.73). The model therefore not only identifies the causal evidence in a scene but also translates it into an appropriate driving decision. The general-purpose VLMs reach at most 4.01 on this overall accuracy, while InternVL3.5-8B produces invalid or unparsable responses for all three CoC metrics. Such margins indicate that causal grounding remains weak in both general-purpose and driving-oriented models, and that targeted CoC supervision substantially narrows this gap. It is worth noting that Alpamayo-1.5 is trained on large-scale proprietary CoC traces yet underperforms on this benchmark. Its lower score under our protocol may reflect instruction-following errors and differences in input format, and should not be interpreted as evidence of weak causal reasoning. By contrast, \ours{}-SFT also performs well on the in-house benchmark of Chinese urban driving decisions. Our training mixture contains no comparable data and this benchmark additionally requires responses in Chinese, which suggests that its decision competence is not confined to the training distribution.

Furthermore, several methods produce invalid or unparsable responses because they cannot follow the required JSON format, while the video-dominated post-training of the Cosmos models may contribute to their weaker results on static single-frame and multi-view inputs. By retaining instruction following and training across diverse input configurations, \ours{}-SFT improves nearly every driving metric without exhibiting these failure modes.

\paragraph{Qualitative Comparison.}
Fig.~\ref{fig:drivevqa_vis} compares four complementary capabilities. In (a), \ours{}-SFT respects the requested temporal scope and reports no parked vehicles in the final frame, whereas Qwen3.5-4B and MiMo-Embodied-7B count agents merely stopped in traffic, conflating stationary with parked. In (b), \ours{}-SFT attributes the stopping decision to the stop sign, while Cosmos-Reason2-32B and Alpamayo-1.5-10B predict continued straight driving based on elements that do not govern the decision. In (c), \ours{}-SFT decomposes the cross-view distance into street width and longitudinal offset, uses parked-vehicle spacing as an implicit scale reference, and predicts $22\,\mathrm{m}$ against a ground truth of $22.93\,\mathrm{m}$. UniDriveVLA-8B instead cites an existing annotation rather than visible scale cues. In (d), \ours{}-SFT identifies the ego lane from the persistent left-turn arrow, whereas Alpamayo-1.5-10B emits a DriveLM-style object reference outside the option set. Cases (c) and (d) also expose differences in instruction following: UniDriveVLA-8B omits the required \texttt{\textbackslash boxed\{\}} delimiter, while \ours{}-SFT adheres to both the format and the answer set. Together, these examples show that \ours{}-SFT integrates evidence across views and time into the requested decision while respecting output constraints. Further cases appear in Appendix~\ref{app:additional_vis}.

\subsubsection{General Vision-Language Understanding}

\begin{table*}[t]
\centering
\caption{
    General vision-language results on (a) knowledge, reasoning, and recognition and (b) spatial understanding and grounding. Avg.\ is the average over the benchmarks. Compared with Qwen3.5-4B, \ours{}-SFT stays within one point on group (a) and surpasses it on group (b), preserving general knowledge and capabilities after driving adaptation. ``--'' indicates an invalid response and is counted as zero in the averages. \textbf{Bold} and \underline{underlined} denote the best and second-best.
}
\label{tab:general_vqa}
\footnotesize
\setlength{\tabcolsep}{3.2pt}

{\textbf{(a) Knowledge, reasoning and recognition}}
\begin{tabular}{l ccccccccccc}
\toprule
\multirow{2.3}{*}{Method} & \multirow{2.3}{*}{\makecell[c]{MM\\Bench}} & \multirow{2.3}{*}{\makecell[c]{MM\\Star}} & \multirow{2.3}{*}{MMMU} & \multicolumn{2}{c}{MMMU-Pro} & \multirow{2.3}{*}{CharXiv} & \multirow{2.3}{*}{\makecell[c]{OCR\\Bench}} & \multirow{2.3}{*}{\makecell[c]{RealWorld\\QA}} & \multirow{2.3}{*}{\makecell[c]{Simple\\VQA}} & \multirow{2.3}{*}{\makecell[c]{Count\\QA}} & \multirow{2.3}{*}{Avg.} \\
\cmidrule(lr){5-6}
& & & & Std & Vis & & & & & & \\
\midrule
InternVL3.5-8B-Inst. & 80.03 & 64.13 & 62.00 & 46.42 & 42.25 & 41.70 & 83.20 & 66.93 & 40.77 & 20.94 & \cellcolor{tabhl}54.84 \\
LLaVA-OV2-8B & 82.66 & 64.93 & 54.67 & 36.30 & 25.95 & 40.10 & 79.30 & 71.76 & 36.68 & 22.58 & \cellcolor{tabhl}51.49 \\
Gemma4-12B & 85.53 & 72.33 & 69.56 & 59.94 & 49.25 & 64.10 & 77.80 & 68.89 & 40.82 & \textbf{39.46} & \cellcolor{tabhl}62.77 \\
Qwen3.5-4B & \underline{87.07} & \underline{75.33} & \textbf{73.44} & \textbf{64.86} & \textbf{61.27} & \textbf{65.10} & 86.90 & \underline{76.34} & \underline{47.84} & \underline{35.86} & \cellcolor{tabhl}\textbf{67.40} \\
\midrule
Cosmos-Reason1-7B & 79.95 & 63.53 & 54.22 & 38.38 & 35.78 & 39.70 & 85.20 & 67.45 & 44.98 & 18.52 & \cellcolor{tabhl}52.77 \\
Cosmos-Reason2-2B & 75.00 & 53.13 & 51.56 & 35.09 & 30.35 & 28.50 & 79.60 & 60.52 & 36.94 & 18.00 & \cellcolor{tabhl}46.87 \\
Cosmos-Reason2-8B & 82.82 & 65.27 & 59.11 & 36.07 & 43.53 & 42.50 & \underline{87.00} & 67.45 & 45.25 & 22.32 & \cellcolor{tabhl}55.13 \\
Cosmos-Reason2-32B & \textbf{88.70} & 72.47 & 61.67 & 41.45 & 52.77 & 53.20 & \textbf{88.20} & 75.69 & \textbf{48.44} & 26.70 & \cellcolor{tabhl}60.93 \\
Cosmos3-nano & 79.57 & 66.67 & 60.89 & 46.36 & 40.75 & 42.10 & 85.20 & 69.67 & 44.99 & 23.63 & \cellcolor{tabhl}55.98 \\
MiMo-Embodied-7B & -- & 22.40 & -- & 27.40 & 28.09 & 57.50 & 78.80 & 28.50 & -- & 22.64 & \cellcolor{tabhl}26.53 \\
UniDriveVLA-8B & 74.30 & 64.07 & 50.67 & 32.43 & 31.56 & 33.80 & 80.20 & 68.10 & 35.26 & 16.88 & \cellcolor{tabhl}48.73 \\
Alpamayo-1.5-10B & 7.51 & 26.13 & 27.44 & 15.61 & 13.47 & 1.50 & 3.20 & 46.93 & -- & 4.71 & \cellcolor{tabhl}14.65 \\
\midrule
\textbf{\ours{}-SFT} & 85.53 & \textbf{75.87} & \underline{72.67} & \underline{62.72} & \underline{59.71} & \underline{64.40} & 86.40 & \textbf{78.95} & 46.12 & 31.74 & \cellcolor{tabhl}\underline{66.41} \\
\bottomrule
\end{tabular}

\vspace{4pt}

\setlength{\tabcolsep}{11.5pt}
{\textbf{(b) Spatial understanding and grounding}}
\begin{tabular}{l ccccc c}
\toprule
Method & EmbSpatial & ERQA & RefSpatial & Omni3D & ODinW13 & Avg. \\
\midrule
InternVL3.5-8B-Inst. & 74.20 & 42.00 & -- & -- & -- & \cellcolor{tabhl}23.24 \\
LLaVA-OV2-8B & 78.43 & 42.25 & -- & -- & -- & \cellcolor{tabhl}24.14 \\
Gemma4-12B & 73.16 & 42.00 & -- & -- & -- & \cellcolor{tabhl}23.03 \\
Qwen3.5-4B & 75.99 & \underline{46.25} & \underline{54.51} & \textbf{47.40} & \underline{40.78} & \cellcolor{tabhl}\underline{52.99} \\
\midrule
Cosmos-Reason1-7B & 68.76 & 38.50 & 0.36 & -- & 4.77 & \cellcolor{tabhl}22.48 \\
Cosmos-Reason2-2B & 66.40 & 38.75 & 32.49 & 31.41 & 33.04 & \cellcolor{tabhl}40.42 \\
Cosmos-Reason2-8B & 77.61 & 43.25 & 51.81 & 32.85 & 40.19 & \cellcolor{tabhl}49.14 \\
Cosmos-Reason2-32B & \textbf{79.26} & 45.25 & \textbf{57.76} & 31.70 & 28.47 & \cellcolor{tabhl}48.49 \\
Cosmos3-nano & 77.88 & 41.25 & -- & 32.26 & 35.87 & \cellcolor{tabhl}37.45 \\
MiMo-Embodied-7B & 45.05 & 39.75 & 2.17 & -- & -- & \cellcolor{tabhl}17.39 \\
UniDriveVLA-8B & 68.16 & 38.00 & 1.44 & 0.33 & -- & \cellcolor{tabhl}21.59 \\
Alpamayo-1.5-10B & 20.58 & 27.50 & -- & -- & -- & \cellcolor{tabhl}9.62 \\
\midrule
\textbf{\ours{}-SFT} & \underline{78.85} & \textbf{48.50} & 50.78 & \underline{45.79} & \textbf{45.87} & \cellcolor{tabhl}\textbf{53.96} \\
\bottomrule
\end{tabular}
\end{table*}

\paragraph{Benchmarks.}
A driving foundation model should acquire domain competence without surrendering the broad visual and world knowledge needed for open-set reasoning and for cockpit applications on an integrated platform. We therefore evaluate fourteen public benchmarks in the two groups of Tab.~\ref{tab:general_vqa}. Group (a) covers knowledge and reasoning with MMBench~\citep{liu2024mmbench}, MMStar~\citep{chen2024we}, MMMU~\citep{yue2024mmmu}, standard and vision splits of MMMU-Pro~\citep{yue2024mmmupro}, and CharXiv~\citep{wang2024charxiv}, together with general recognition through OCRBench~\citep{liu2024ocrbench}, RealWorldQA~\citep{xai2024grok}, SimpleVQA~\citep{cheng2025simplevqa} for multimodal factuality, and CountQA~\citep{tamarapalli2025countqa} for object counting. Group (b) covers spatial understanding and grounding with EmbSpatial-Bench~\citep{du2024embspatial}, ERQA~\citep{team2025gemini}, RefSpatial-Bench~\citep{zhou2025roborefer}, Omni3D-Bench~\citep{marsili2025visual}, and ODinW13~\citep{li2022grounded}.

\paragraph{Results.}
As shown in Tab.~\ref{tab:general_vqa}, \ours{}-SFT preserves broad vision-language competence after adaptation on large-scale driving data. On the knowledge, reasoning, and recognition benchmarks in Tab.~\ref{tab:general_vqa}(a), it averages 66.41 versus 67.40 for Qwen3.5-4B, staying within one point while ranking first or second on 6 of the 10 settings, including the best scores on MMStar and RealWorldQA. On the spatial understanding and grounding benchmarks in Tab.~\ref{tab:general_vqa}(b), it averages 53.96 and even exceeds Qwen3.5-4B at 52.99, with the best scores on ERQA and ODinW13 covering embodied spatial reasoning and open-set object grounding. Driving adaptation therefore leaves knowledge-intensive capabilities essentially intact and strengthens spatially grounded ones.

This preservation distinguishes \ours{}-SFT from comparison methods specialized for physical AI, embodied reasoning, and autonomous driving. It matches or exceeds Cosmos-Reason2-32B on 10 of the 15 settings and achieves a 5.48-point higher average across them. Several other specialized models produce invalid responses on some benchmarks, showing that strong domain specialization does not necessarily preserve the general instruction-following interface required across diverse tasks. In contrast, \ours{}-SFT improves driving understanding while maintaining both this interface and broad visual and world knowledge. Together, these results suggest the potential for more reliable generalization to rare and previously unseen situations in open-world driving.

The preserved capabilities also matter for deployment. Under cockpit-driving integration, a single onboard model is expected to serve cockpit applications such as dialogue and open-ended visual queries, alongside driving perception and planning. \ours{}-SFT loses less than one point on group (a) and gains on group (b) while acquiring driving competence, so one instance can cover both domains. This removes the need for a separate cockpit model and the associated compute and maintenance cost. Methods that lose general capability during driving adaptation cannot provide this saving, even when their driving scores are competitive.

\subsection{Motion Planning}
\label{sec:exp_planning}

\paragraph{Benchmarks.}
To evaluate motion planning at progressively greater levels of interaction, we consider two open-loop benchmarks, one pseudo-closed-loop benchmark, and one closed-loop simulator.

\textbf{1) WOD-E2E}~\citep{xu2026wod} focuses on challenging long-tail scenarios and provides several human-rated candidate trajectories for each scene. Its primary metric is the Rater Feedback Score (RFS), which matches a prediction to these candidates under a speed-dependent tolerance and assigns the rating of the closest valid match. RFS can therefore recognize an acceptable plan even when it differs from the recorded future. We also report average displacement error (ADE), defined as the mean Euclidean position error relative to the recorded trajectory, at 3\,s and 5\,s.

\textbf{2) PAI-AV}~\citep{nvidia2025physicalai} requires each method to predict six candidate trajectories per scene. We report their average ADE to measure overall trajectory quality and the minimum ADE (minADE) to determine whether any candidate covers the recorded motion. We evaluate both metrics at 3\,s and 5\,s. The standard 644-example split overlaps with publicly available training data by official construction. To ensure a fair comparison, we also report results on a leakage-free subset curated from held-out test clips.

\textbf{3) NAVSIM}~\citep{dauner2024navsim} follows a pseudo-closed-loop protocol. The planner is queried only once, and surrounding agents replay their recorded motion without reacting to the ego vehicle. On NAVSIM v1.1 navtest, we report PDMS and its five components. No collision (NC) and drivable area compliance (DAC) act as multiplicative factors on a weighted combination of ego progress (EP), time-to-collision (TTC), and comfort (Comf.).

\textbf{4) AlpaSim}~\citep{alpasim_2025} evaluates closed-loop behavior under compounding errors. We use PAI-AV-NuRec~\citep{wu20253dgut} version 26.02 to simulate 916 scenarios with novel views as the ego vehicle deviates from recorded logs. Following Alpamayo~\citep{wang2025alpamayo}, we measure close encounter rate (all-event and at-fault), off-road rate, progress, and AlpaSim score (all-event and at-fault). These metrics quantify close-proximity events, roadway departures, scenario completion, and average distance traveled between events, respectively. The \emph{at-fault} variants exclude events not caused by the ego vehicle.

\begin{table*}[!t]
\caption{Open-loop planning on WOD-E2E. We report average displacement error (ADE) at 3\,s and 5\,s and the Rater Feedback Score (RFS). RL indicates reinforcement learning. Because the validation split is used for RL, the test split provides a more informative assessment of generalization.}
\label{tab:waymo_e2e}
\footnotesize
\centering
\begin{subtable}[t]{0.49\textwidth}
\centering
\caption{Validation split.}
\label{tab:waymo_e2e_val}
\renewcommand{\arraystretch}{1.62}
\fontsize{7.6}{9.1}\selectfont
\setlength{\tabcolsep}{0.9pt}
\begin{tabular*}{\linewidth}{@{\extracolsep{\fill}}lcccc@{}}
\toprule
Method & RL & \makecell{ADE$\downarrow$\\3\,s} & \makecell{ADE$\downarrow$\\5\,s} & RFS$\uparrow$ \\
\midrule
Human Driver~\citep{rowe2025poutine} & -- & -- & -- & 8.13 \\
VAD~\citep{jiang2023vad} &-- & 3.19 & 5.81 & 4.45 \\
UniAD~\citep{hu2023planning} &-- & 6.50 & 10.81 & 5.78 \\
RAP-DINO~\citep{feng2026rap} &-- & 0.97 & 2.20 & 7.91 \\
MindVLA-U1~\citep{huang2026mindvla} &-- & 0.89 & 2.11 & 7.92 \\
MindVLA-U1~\citep{huang2026mindvla} & \checkmark & 1.01 & 2.28 & 8.20 \\
\midrule
\textbf{\ours{}-SFT} w/o reasoning &-- & 0.99 & 2.33 & 7.95 \\
\textbf{\ours{}-SFT} w/ reasoning & -- & 0.99 &2.31 & 7.95 \\
\textbf{\ours{}-RL} & \checkmark & 0.62 & 1.27 & 8.45 \\
\bottomrule
\end{tabular*}
\end{subtable}
\hfill
\begin{subtable}[t]{0.495\textwidth}
\centering
\caption{Test split.}
\label{tab:waymo_e2e_test}
\fontsize{7.6}{9.1}\selectfont
\setlength{\tabcolsep}{0.9pt}
\begin{tabular*}{\linewidth}{@{\extracolsep{\fill}}lcccc@{}}
\toprule
Method & RL & \makecell{ADE$\downarrow$\\3\,s} & \makecell{ADE$\downarrow$\\5\,s} & RFS$\uparrow$ \\
\midrule
Swin-Trajectory~\citep{park2025swin} & -- & 1.21 & 2.81 & 7.54 \\
DiffusionLTF~\citep{nguyen2025open} & -- & 1.36 & 2.89 & 7.72 \\
UniPlan~\citep{liao2025diffusiondrive} &-- & 1.31 & 2.99 & 7.78 \\
\midrule
LightEMMA~\citep{qiao2025lightemma} &-- & 1.71 & 3.74 & 6.52 \\
NaiveEMMA~\citep{xu2026wod} & -- & 1.32 & 3.02 & 7.53 \\
dVLM-AD~\citep{ma2026dvlm} & -- & 1.29 & 3.02 & 7.63 \\
HMVLM~\citep{wang2025hmvlm} & -- & 1.33 & 3.07 & 7.74 \\
MindVLA-U1~\citep{huang2026mindvla} & -- & 1.16 & 2.67 & 7.77 \\
AutoVLA~\citep{zhou2026autovla} & \checkmark & 1.35 & 2.96 & 7.56 \\
NoRD~\citep{rawal2026nord} & \checkmark & 1.25 & -- & 7.71 \\
MindVLA-U1~\citep{huang2026mindvla} & \checkmark & 1.09 & 2.66 & 7.87 \\
\midrule
\textbf{\ours{}-SFT} w/o reasoning & -- & 1.20 & 2.66 & 7.76 \\
\textbf{\ours{}-SFT} w/ reasoning & -- & 1.19 & 2.65 & 7.78 \\
\textbf{\ours{}-RL} & \checkmark & 1.19 & 2.67 & 7.91 \\
\bottomrule
\end{tabular*}
\end{subtable}
\end{table*}

\paragraph{Results.}
We evaluate \ours{}-SFT with and without the planning-reasoning condition, together with \ours{}-RL after reinforcement learning. Using 2.83M training samples assembled from public sources, our method attains the highest PDMS on NAVSIM and the highest test-split RFS on WOD-E2E among the compared methods. We analyze the results at increasing levels of interaction, from open-loop prediction to pseudo-closed-loop scoring and closed-loop simulation.

\textbf{Open-loop.} On the long-tail WOD-E2E test split in Tab.~\ref{tab:waymo_e2e_test}, \ours{}-SFT with reasoning achieves an RFS of 7.78, slightly exceeding MindVLA-U1 at 7.77 before reinforcement learning. Reasoning raises RFS from 7.76 to 7.78, even though only 142K of the 557K WOD-E2E samples provide reasoning-conditioned supervision. On the validation split in Tab.~\ref{tab:waymo_e2e_val}, which supplies the RFS annotations used for reinforcement learning, \ours{}-RL improves RFS from 7.95 to 8.45 while significantly reducing the 3\,s and 5\,s ADE. The resulting RFS exceeds the human-driver reference of 8.13. Since these annotations supervise the reward, this in-sample result indicates effective optimization of preference alignment on the training scenarios rather than generalization beyond human driving. More importantly, the improvement transfers to the test split. Reinforcement learning raises RFS from 7.78 to 7.91, exceeding the reinforced MindVLA-U1 by 0.04 points. The held-out gain indicates improved alignment with human preference without materially changing displacement from the recorded future.

\begin{table*}[!t]
\caption{Open-loop motion planning on PAI-AV. We report the average and minimum ADE (m) over six trajectories at 3\,s and 5\,s on the standard 644-example split and a leakage-free 700-frame subset curated from held-out test clips. All comparison methods are reproduced under the same evaluation setting.}
\label{tab:pai_av_openloop}
\footnotesize
\centering
\setlength{\tabcolsep}{9.7pt}
\begin{tabular}{l cccc cccc}
\toprule
\multirow{3.6}{*}{Method} & \multicolumn{4}{c}{644-example split} & \multicolumn{4}{c}{700-frame subset} \\
\cmidrule(lr){2-5} \cmidrule(lr){6-9}
& \multicolumn{2}{c}{Avg.\ ADE$\downarrow$} & \multicolumn{2}{c}{minADE$\downarrow$} & \multicolumn{2}{c}{Avg.\ ADE$\downarrow$} & \multicolumn{2}{c}{minADE$\downarrow$} \\
\cmidrule(lr){2-3} \cmidrule(lr){4-5} \cmidrule(lr){6-7} \cmidrule(lr){8-9}
& 3\,s & 5\,s & 3\,s & 5\,s & 3\,s & 5\,s & 3\,s & 5\,s \\
\midrule
Alpamayo-R1-10B~\citep{wang2025alpamayo} & 0.37 & 1.13 & 0.16 & 0.48 & 0.41 & 1.22 & 0.18 & 0.51 \\
Alpamayo-1.5-10B~\citep{wang2025alpamayo} & 0.35 & 1.05 & 0.16 & 0.50 & 0.36 & 1.06 & 0.17 & 0.49 \\
DriveWAM~\citep{shi2026drivewam} & 0.67 & -- & 0.37 & -- & 0.69 & -- & 0.38 & -- \\
SimWAM~\citep{zhao2026simwam} &0.41 &-- &0.38 & -- &0.43 &-- & 0.40 & -- \\
\midrule
\textbf{\ours{}-SFT} w/o reasoning & 0.38 & 1.07 & 0.34 & 0.96 & 0.43 & 1.24 & 0.39 & 1.11 \\
\textbf{\ours{}-SFT} w/ reasoning & 0.37 & 1.07 & 0.34 & 0.97 & 0.42 & 1.23 & 0.39 & 1.11 \\
\textbf{\ours{}-RL} & 0.42 & 1.11 & 0.38 & 1.00 & 0.47 & 1.27 & 0.43 & 1.15 \\
\bottomrule
\end{tabular}
\end{table*}

\begin{table}[!t]
  \caption{Pseudo-closed-loop motion planning on NAVSIM v1.1 \texttt{navtest}. We report the Predictive Driver Model Score (PDMS) and its no-collision (NC), drivable area compliance (DAC), ego progress (EP), time-to-collision (TTC), and comfort (Comf.) components. RL indicates reinforcement learning. $\ddagger$ denotes best-of-$N$ selection with $N=6$, where the candidate with the highest PDMS is chosen for each scene.}
  \label{tab:navsim_v1}
  \centering
  \footnotesize
  \setlength{\tabcolsep}{8.4pt}
  \begin{tabular}{lccccccc}
    \toprule
    Method & RL & NC $\uparrow$ & DAC $\uparrow$ & EP $\uparrow$ & TTC $\uparrow$ & Comf. $\uparrow$ & PDMS $\uparrow$ \\
    \midrule
    TransFuser \citep{chitta2022transfuser} & - & 97.7 & 92.8 & 79.2 & 92.8 & 100.0 & 84.0 \\
    DRAMA \citep{yuan2024drama} & - & 98.0 & 93.1 & 80.1 & 94.8 & 100.0 & 85.5 \\
    Hydra-MDP \citep{li2024hydra} & - & 98.3 & 96.0 & 78.7 & 94.6 & 100.0 & 86.5 \\
    DiffusionDrive \citep{liao2025diffusiondrive} & - & 98.2 & 96.2 & 82.2 & 94.7 & 100.0 & 88.1 \\
    \midrule
    Epona \citep{zhang2025epona} & - & 97.9 & 95.1 & 80.4 & 93.8 & 99.9 & 86.2 \\
    ReCogDrive \citep{li2025recogdrive} & - & 98.3 & 95.1 & 81.1  & 94.3 & 100.0 & 86.8 \\
    AutoVLA \citep{zhou2026autovla} &- & 96.9 & 92.4 & 75.8 & 88.1 & 99.9 & 80.5 \\
    SpanVLA~\citep{zhou2026spanvla} & - & 97.5 & 90.8 & 76.9 & 93.7 & 99.5 & 82.1 \\
    \midrule
    \textbf{\ours{}-SFT} w/o reasoning & - & 98.2 & 96.4 & 82.0 & 94.4 & 100.0 & 87.8 \\
    \textbf{\ours{}-SFT} w/ reasoning & - & 98.4 & 96.6 & 82.4 & 94.7 & 100.0 & 88.2 \\
    \textbf{\ours{}-SFT}$^\ddagger$ w/o reasoning & - & 98.6 & 97.1 & 82.9 & 95.1 & 100.0 & 88.9 \\
    \textbf{\ours{}-SFT}$^\ddagger$ w/ reasoning & - & 98.7 & 97.2 & 83.2 & 95.5 & 100.0 & 89.3 \\
    \midrule
    \midrule
    ReCogDrive \citep{li2025recogdrive} & \checkmark & 98.2 & 97.8 & 83.5 & 95.2 & 99.8 & 89.6 \\
    AutoVLA \citep{zhou2026autovla} & \checkmark & 98.4 & 95.6 & 81.9 & 98.0 & 99.9 & 89.1 \\
    SpanVLA~\citep{zhou2026spanvla} & \checkmark & 99.1 & 97.1 & 86.3 & 95.2 & 100.0 & 90.3 \\
    ExploreVLA~\citep{sheng2026explorevla} & \checkmark & 98.8 & 98.4 & 83.5 & 96.5 & 99.9 & 90.4 \\
    EponaV2~\citep{xu2026eponav2} &\checkmark & 98.6 & 97.9 & 84.8 & 95.7 & 100.0 & 90.4 \\
    \midrule
    \textbf{\ours{}-RL} & \checkmark & 98.6 & 98.2 & 84.8 & 95.9 & 100.0 & 90.7 \\
    \textbf{\ours{}-RL$^\ddagger$} & \checkmark & 98.8 & 98.4 & 85.5 & 96.5 & 100.0 & 91.4 \\
    \bottomrule
  \end{tabular}
\end{table}

Complementing the preference-based evaluation on WOD-E2E, PAI-AV assesses both the quality and diversity of six predicted trajectories. We reproduce all comparison methods under the same evaluation setting. As shown in Tab.~\ref{tab:pai_av_openloop}, on the leakage-free subset, \ours{}-SFT with reasoning attains a 3\,s average ADE of 0.42\,m, compared with 0.36\,m for Alpamayo-1.5. Our minADE is 0.39\,m versus 0.17\,m, and the narrow gap between average ADE and minADE suggests that our candidates remain concentrated around similar motions. Part of this difference is plausibly attributable to training scale. Alpamayo-1.5 uses 80,000 hours of driving trajectories and 3M CoC reasoning traces, whereas PAI-AV contains 156K clips, corresponding to approximately 900 raw hours before sparse frame sampling. DriveWAM and SimWAM both incorporate future generative supervision, yet exhibit markedly different candidate distributions. DriveWAM achieves a much lower minADE than average ADE, suggesting broader candidate coverage but weaker typical trajectory accuracy. SimWAM substantially improves average ADE to 0.43\,m, while its minADE remains close at 0.40\,m, indicating more limited diversity. In comparison, \ours{}-SFT achieves a slightly lower average ADE of 0.42\,m with a comparable minADE of 0.39\,m. Reinforcement learning increases the PAI-AV errors by only 3 to 5\,cm. This modest open-loop trade-off accompanies improved preference alignment on WOD-E2E, higher pseudo-closed-loop PDMS on NAVSIM, and a halving of the closed-loop off-road rate in AlpaSim.

\begin{table*}[!t]
\caption{Closed-loop planning on 916 AlpaSim~\citep{alpasim_2025} scenarios using PAI-AV-NuRec~\citep{wu20253dgut} version 26.02. Params. denotes all parameters excluding the LLM token embeddings. All comparison methods are reproduced under the same evaluation setting.}
\label{tab:alpasim_closedloop}
\footnotesize
\centering
\setlength{\tabcolsep}{1.2pt}
\begin{tabular}{l ccc c cc c}
\toprule
\multirow{2.3}{*}{Method}& \multirow{2.3}{*}{Params.} & \multicolumn{2}{c}{Close Encounter Rate (\%)$\downarrow$} & \multirow{2.3}{*}{\makecell{Off-Road\\Rate (\%)$\downarrow$}} & \multirow{2.3}{*}{Progress (\%)$\uparrow$} & \multicolumn{2}{c}{AlpaSim Score$\uparrow$}\\
\cmidrule(lr){3-4} \cmidrule(lr){7-8}
&& all & at-fault & & & all & at-fault \\
\midrule
Alpamayo-R1~\citep{wang2025alpamayo} & 9.8 B & 19.0 & 6.0 & 17.0 & 67.0 & 0.36 & 0.58 \\
Alpamayo-1.5~\citep{wang2025alpamayo} & 9.8 B & 37.0 & 11.0 & 16.0 & 59.0 & 0.23 & 0.45 \\
DriveWAM~\citep{shi2026drivewam} & 15.0 B & 56.0  & 5.0 & 8.0 & 35.0 & 0.10 & 0.53 \\
SimWAM~\citep{zhao2026simwam} & 6.0 B &35.0 &22.0 &19.0 &62.0 & 0.22 & 0.30 \\
\midrule
\textbf{\ours{}-SFT} w/ reasoning & 5.0 B & 38.0 & 12.0 & 24.0 & 54.0 & 0.16 & 0.27 \\
\textbf{\ours{}-RL} & 5.0 B & 41.0 & 11.0 & 12.0 & 48.0 & 0.16 & 0.37 \\
\bottomrule
\end{tabular}
\end{table*}

\textbf{Pseudo-closed-loop.} NAVSIM advances one step toward closed-loop evaluation by propagating the predicted trajectory through a vehicle model. However, it does not re-query the planner and keeps surrounding agents non-reactive, so it cannot capture compounding errors. As shown in Tab.~\ref{tab:navsim_v1}, before reinforcement learning, \ours{}-SFT with reasoning reaches 88.2 PDMS and surpasses comparison methods trained only with imitation, including DiffusionDrive at 88.1. This competitive result may draw on capabilities acquired during vision-language pretraining, together with the scale and behavioral diversity of the unified planning data. Reasoning improves PDMS by 0.4 points despite being available for only 78K NAVSIM samples, indicating that the textual condition provides useful context for trajectory generation. Reinforcement learning yields a further 2.5-point gain, raising PDMS to a promising 90.7. These results suggest that reinforcement on NAVSIM is particularly effective at reducing conservative, low-progress behavior. Following a common practice of reporting the best among multiple sampled trajectories, we also select per scene the candidate with the highest PDMS among six samples, which raises PDMS to 91.4 for \ours{}-RL and 89.3 for \ours{}-SFT with reasoning. This margin over the single-trajectory output indicates that the sampled set already contains stronger trajectories than the default prediction, suggesting headroom that better inference-time selection could recover. Nevertheless, PDMS should not be treated as a direct proxy for interactive driving quality. Near the upper end of this benchmark, further gains may increasingly reflect adaptation to the scoring function, while the non-reactive protocol cannot reveal how errors accumulate during interaction.

\textbf{Closed-loop.} AlpaSim completes this progression by repeatedly querying the planner as its actions alter subsequent observations, thereby exposing compounding errors and recovery behavior that NAVSIM cannot measure. For a controlled comparison, we reproduce Alpamayo-R1 and Alpamayo-1.5 under the same evaluation setting. As shown in Tab.~\ref{tab:alpasim_closedloop}, \ours{}-RL achieves an at-fault close encounter rate of 11.0\%, matching Alpamayo-1.5, while their all-event rates are 41.0\% and 37.0\%, respectively. Reinforcement learning halves the off-road rate from 24.0\% to 12.0\%, a rate lower than that of either Alpamayo variant, and raises the at-fault AlpaSim score from 0.27 to 0.37. These safety gains come with lower progress, which decreases from 54.0\% to 48.0\%, and a modest increase in the all-event close encounter rate. The results indicate a shift toward safer but more conservative behavior. Despite matching Alpamayo-1.5 in at-fault close encounter rate, \ours{}-RL records lower all-event and at-fault AlpaSim scores of 0.16 and 0.37, compared with 0.23 and 0.45. One possible factor is the temporal sampling of the visual input. AlpaSim repeatedly replans over short intervals and therefore emphasizes rapid responses to recent visual changes. Alpamayo-1.5 observes a dense 0.4\,s visual history, whereas our input contains four observations sampled at 0.5\,s intervals over 1.5\,s. This broader but sparser history may limit responsiveness over the short replanning horizon.

An AlpaSim score should not be interpreted in isolation. DriveWAM~\citep{shi2026drivewam} attains an at-fault score of 0.53, but its progress is only 35.0\%. We find that it often remains stationary or advances only briefly. This behavior reduces ego-at-fault and off-road events, which can raise the at-fault score because the metric divides the traveled distance by the corresponding event count. However, a nearly stationary ego vehicle remains susceptible to interactions caused by following traffic. DriveWAM thus records an all-event close encounter rate of 56.0\% and an all-event score of 0.10. SimWAM~\citep{zhao2026simwam} shows the opposite behavior. We find that its driving policy is considerably more aggressive, frequently accelerating forward while failing to decelerate sufficiently for preceding vehicles or obstacles. This yields relatively high progress of 62.0\%, but also leads to an at-fault close encounter rate of 22.0\% and an off-road rate of 19.0\%, resulting in an at-fault score of only 0.30. In comparison, \ours{}-RL achieves substantially lower at-fault close encounters and off-road violations of 11.0\% and 12.0\%, respectively, while maintaining 48.0\% progress. These results highlight the importance of jointly considering progress, safety events, and AlpaSim scores when assessing effective closed-loop driving.

\begin{figure*}[t]
\centering
\includegraphics[width=\textwidth]{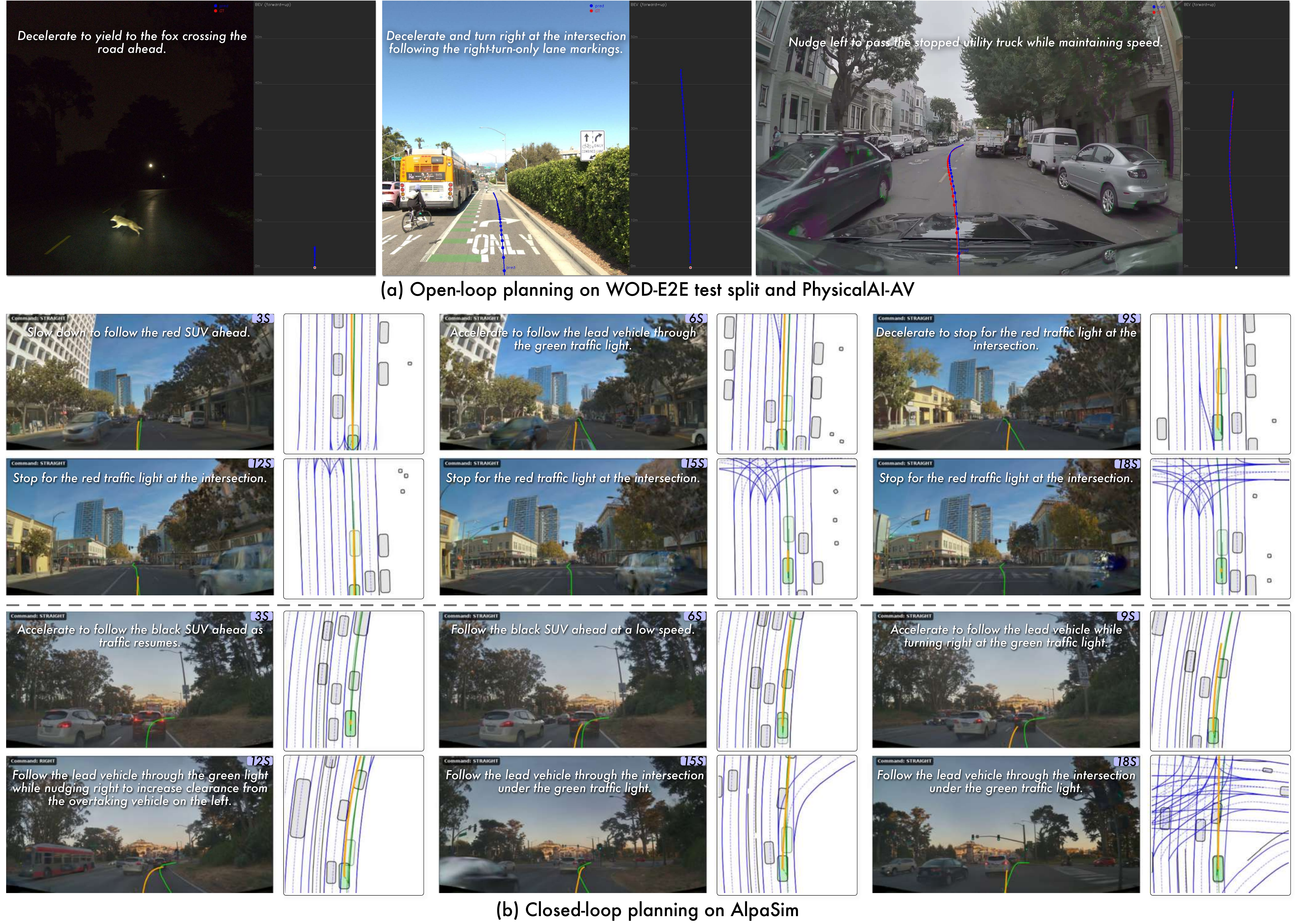}
\caption{
    Qualitative motion planning results. (a) Open-loop predictions from \ours{}-SFT with reasoning on the WOD-E2E test split (left and middle) and PAI-AV (right). WOD-E2E provides no ground-truth trajectory for the test split, while the PAI-AV example shows both the prediction and recorded future. (b) Two closed-loop AlpaSim rollouts from \ours{}-RL at selected timestamps, with the predicted and recorded trajectories shown for comparison.
}
\label{fig:planning_vis}
\end{figure*}

\begin{figure*}[t]
\centering
\includegraphics[width=\textwidth]{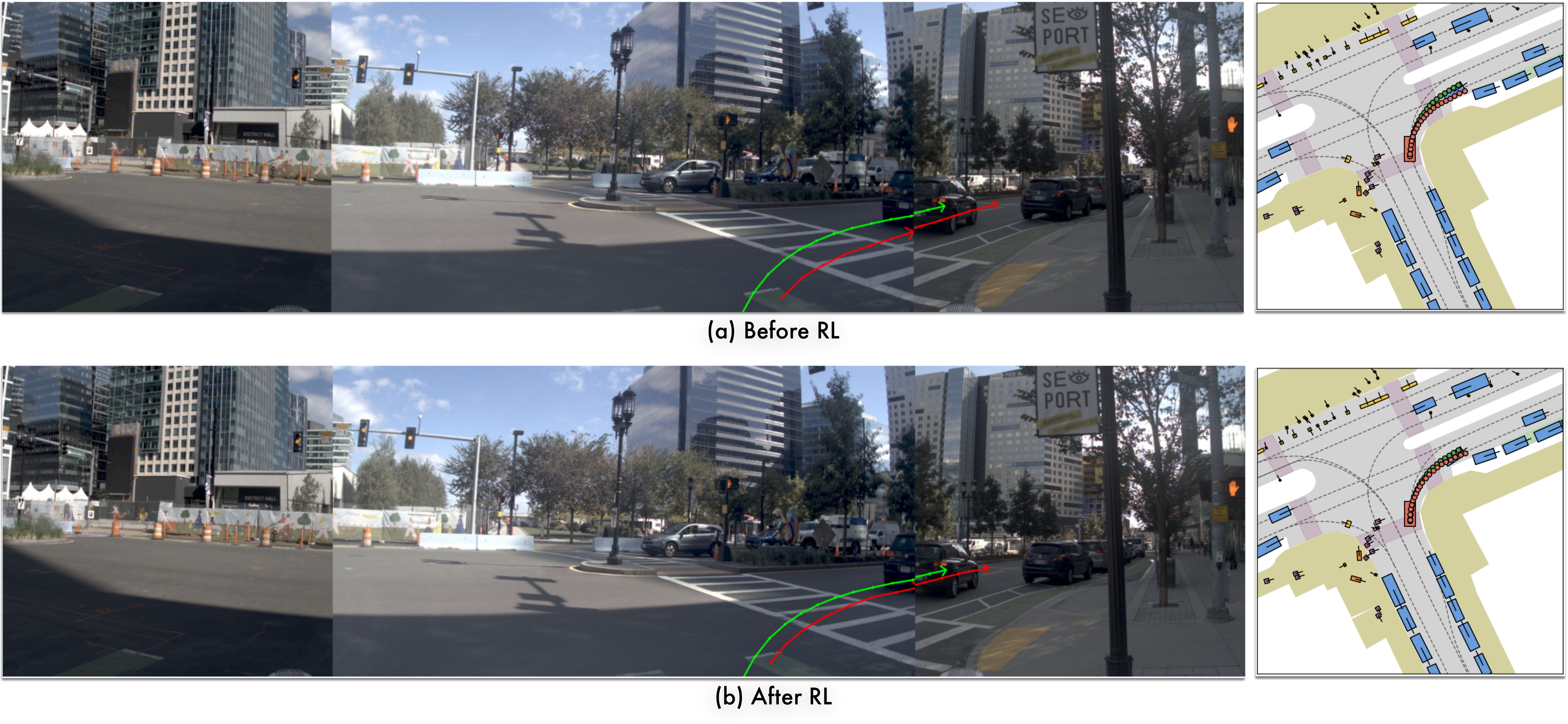}
\caption{
    Qualitative effect of reinforcement learning on the same NAVSIM left-turn scene. (a) \ours{}-SFT with reasoning before reinforcement learning. (b) \ours{}-RL. The prediction is shown in red and the recorded future in green in the camera and BEV views.
}
\label{fig:navsim_rl}
\end{figure*}

\paragraph{Qualitative Results.}
Fig.~\ref{fig:planning_vis} complements the quantitative evaluation with open-loop and closed-loop planning. In Fig.~\ref{fig:planning_vis}(a), \ours{}-SFT with reasoning grounds its plans in relevant scene evidence. It decelerates for a crossing animal, follows the right-turn-only lane, and adjusts laterally to pass a stopped vehicle. On PAI-AV, the predicted trajectory remains close to the recorded future. Fig.~\ref{fig:planning_vis}(b) shows two AlpaSim rollouts from \ours{}-RL. In the first rollout, the model follows the lead vehicle through a green light and subsequently stops when the signal turns red. In the second, it follows a slower vehicle, turns right at the intersection, and adjusts its lateral position as another vehicle overtakes. Across both rollouts, the reasoning and trajectory are updated with the evolving traffic state while remaining consistent with the navigation instruction and road geometry.

Fig.~\ref{fig:navsim_rl} further illustrates how reinforcement learning adapts trajectory predictions to the NAVSIM objective. Both variants retain the same high-level left-turn maneuver, while \ours{}-RL reduces a small lateral deviation from the recorded future. This fine-grained correction better aligns the trajectory with the NAVSIM scoring criteria and is consistent with the PDMS improvement in Tab.~\ref{tab:navsim_v1}.

\subsection{Ablation Study and Analysis}
\label{sec:exp_ablation}

\begin{table}[t]
\centering
\caption{
    Ablation of the Stage 2 training mixture. Row \textsc{i} uses the unadapted Qwen3.5-4B, while rows \textsc{ii} and \textsc{iii} progressively introduce vision-language and 3D perception supervision. For each row, we train a Planning Expert for 15 epochs on the same WOD-E2E data and evaluate RFS on the validation split.
}
\label{tab:ablation_vl}
\footnotesize
\setlength{\tabcolsep}{13pt}
\begin{tabular}{c cc ccc c}
\toprule
\multirow{3.3}{*}{ID} & \multicolumn{2}{c}{Stage 2 Training Mixture} & \multicolumn{3}{c}{Vision-Language Evaluation} & Planning \\
\cmidrule(lr){2-3} \cmidrule(lr){4-6} \cmidrule(lr){7-7}
& \makecell[c]{Vision-\\Language}  & \makecell[c]{3D\\Perception} & \makecell[c]{Driving QA\\Avg.$\uparrow$} & \makecell[c]{CoC\\Overall$\uparrow$} & \makecell[c]{General VQA\\Avg.$\uparrow$} & \makecell[c]{WOD-E2E\\RFS$\uparrow$} \\
\midrule
\textsc{i} & \xmark & \xmark & 63.52 & 2.58 & 62.60 & 7.88 \\
\textsc{ii} & \cmark  & \xmark & 70.07 & 40.97 & 63.18 & 7.91 \\
\textsc{iii} & \cmark & \cmark & 69.43 & 41.26 & 62.26 & 7.96 \\
\bottomrule
\end{tabular}
\end{table}

\begin{figure*}[t]
\centering
\includegraphics[width=\textwidth]{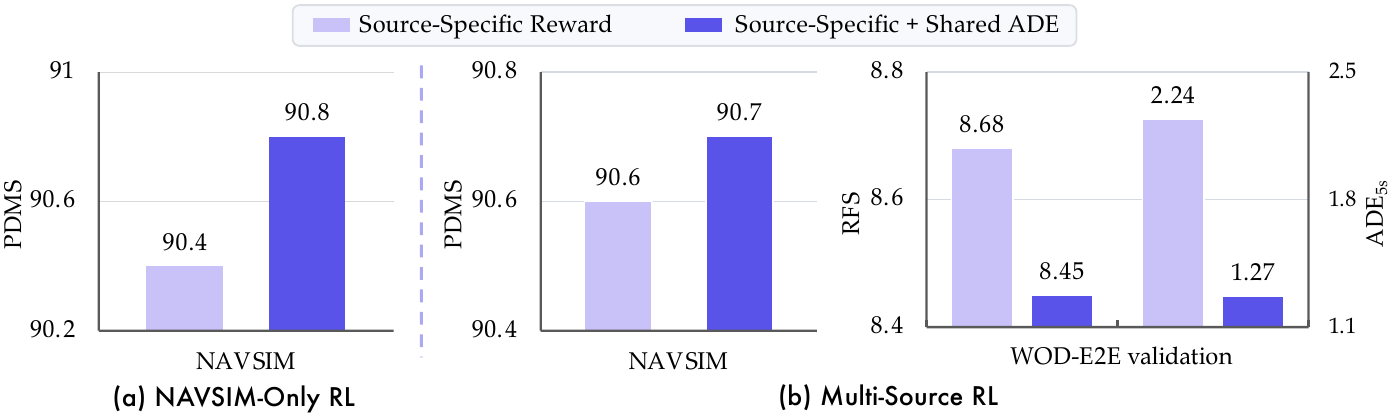}
\caption{
    Ablation of the reinforcement learning data mixture and reward design.
    (a) Reinforcement learning on NAVSIM alone.
    (b) Joint reinforcement learning on NAVSIM, WOD-E2E, and PAI-AV.
    The source-specific reward uses PDMS for NAVSIM, RFS for WOD-E2E, and ADE for PAI-AV.
    The shared-ADE variant adds a displacement reward to every data source.
}
\label{fig:reward_ablation}
\end{figure*}

\paragraph{Analysis of the Stage 2 Training Mixture.}

First, we examine how the Stage 2 training mixture in Sec.~\ref{sec:training} affects vision-language capability, explicit 3D perception, and subsequent planning. Starting from the unadapted Qwen3.5-4B in row \textsc{i}, row \textsc{ii} introduces vision-language training, and row \textsc{iii} additionally enables 3D perception supervision. For a controlled planning comparison, we train a Planning Expert for each variant on WOD-E2E~\citep{xu2026wod} for 15 epochs in Stage 3 and report RFS on the validation split. Tab.~\ref{tab:ablation_vl} shows that vision-language training improves Driving QA Avg.\ by 6.55 points and produces a substantially larger gain in CoC reasoning, while preserving general vision-language capability. Adding 3D perception supervision keeps all three vision-language aggregates within one point of row \textsc{ii}. Consistent with Sec.~\ref{sec:exp_perception}, the BEV perception head serves as a 3D probe by exposing how readily the shared representations support 3D detection, semantic occupancy, and BEV map predictions. Its perception objectives further provide task-specific 3D supervision to the shared visual pathway during joint adaptation. The complete Stage 2 mixture adds an explicit, inspectable 3D perception capability while largely preserving vision-language performance. It also yields the highest RFS after Stage 3, although the margin over the other variants is small. The planning result supports the compatibility of the Stage 2 mixture with subsequent Planning Expert training, but does not establish explicit 3D supervision as the source of the improvement.

\paragraph{Analysis of Reinforcement Rewards.}
We next ablate the data mixture and the shared ADE term used for reinforcement learning, as shown in Fig.~\ref{fig:reward_ablation}. With NAVSIM-only training, augmenting the source-specific PDMS reward with ADE raises PDMS from 90.4 to 90.8. Joint training on NAVSIM, WOD-E2E, and PAI-AV produces similar NAVSIM scores of 90.6 and 90.7 without and with the shared ADE term, respectively. The small differences from single-source training suggest limited cross-dataset interference, particularly when each source provides a task-aligned reward. On the WOD-E2E validation split, adding ADE lowers RFS from 8.68 to 8.45 but substantially reduces the 5\,s ADE from 2.24\,m to 1.27\,m. The shared displacement reward anchors preference optimization to the recorded motion and limits excessive trajectory deviation. Based on this trade-off, our final recipe jointly trains on all three sources and includes the shared ADE term for every source.

\begin{figure*}[t]
\centering
\includegraphics[width=\textwidth]{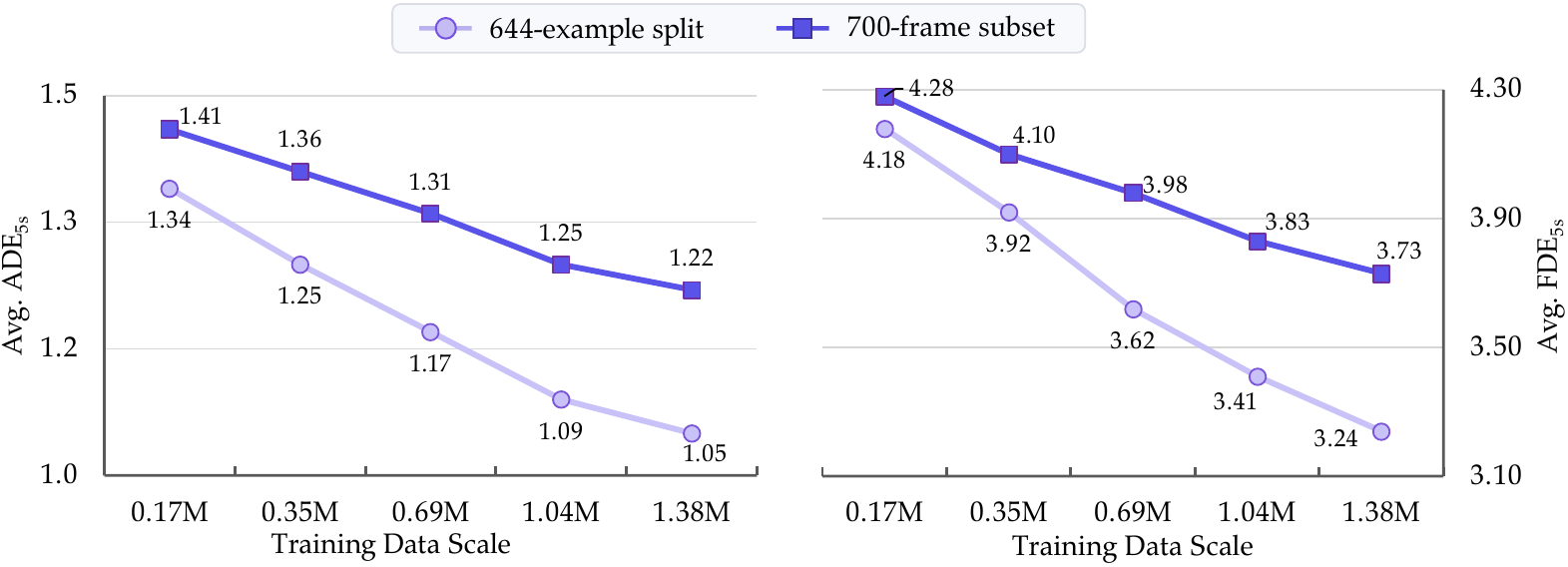}
\caption{
    Effect of PAI-AV planning data scale. The Planning Expert is trained in Stage 3 using only PAI-AV, with the number of training samples increasing from 0.17M to 1.38M. Both 5\,s Avg. ADE and Avg. FDE decrease consistently on the standard 644-example split and the leakage-free 700-frame subset.
}
\label{fig:data_scale_ablation}
\end{figure*}

\begin{figure*}[t]
\centering
\includegraphics[width=\textwidth]{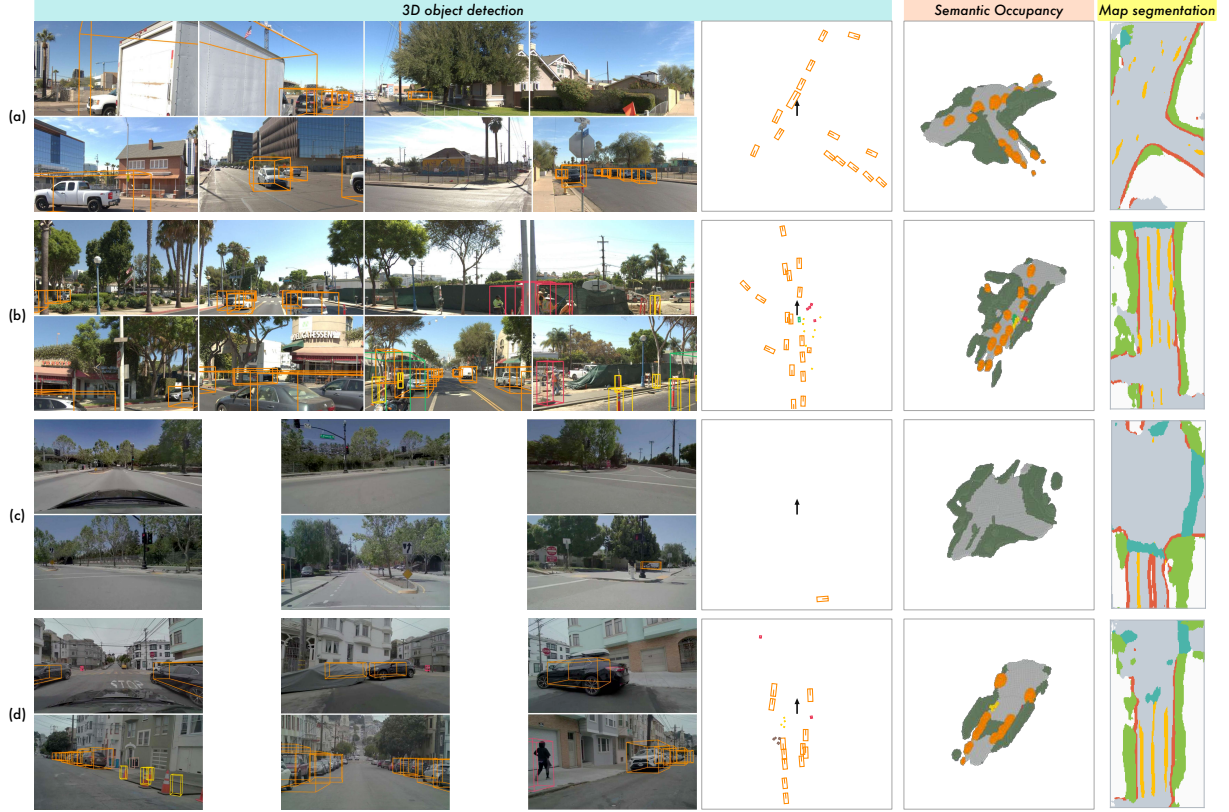}
\caption{
    Qualitative perception outputs of \ours{}-SFT on unseen camera rigs. WOD-E2E is shown in (a, b) and PAI-AV in (c, d). These datasets provide no unified perception ground truth, so all panels show predictions only.
}
\label{fig:ood_perception_vis}
\end{figure*}

\paragraph{Effect of Planning Data Scale.} We further examine how Stage 3 planning performance scales with the amount of trajectory supervision. To isolate the effect of data scale from cross-dataset mixing, we train the Planning Expert exclusively on PAI-AV using 0.17M, 0.35M, 0.69M, 1.04M, and 1.38M training samples. As shown in Fig.~\ref{fig:data_scale_ablation}, both 5\,s Avg. ADE and Avg. FDE decrease monotonically as the training set grows. On the standard split, Avg. ADE decreases from 1.34\,m to 1.05\,m, while Avg. FDE decreases from 4.18\,m to 3.24\,m. The leakage-free subset remains more challenging but follows the same trend, showing that the gains are not confined to the standard split. Performance continues to improve at 1.38M samples, with no clear sign of saturation at the current data scale.

\paragraph{Qualitative Transfer to Unseen Camera Rigs.}
The quantitative comparison in Sec.~\ref{sec:exp_perception} covers only the two sources that supervise perception. We finally examine whether the perception pathway can operate directly on camera configurations never seen in training. To this end, we sample frames from the eight-camera ring rig of WOD-E2E~\citep{xu2026wod} and the six-camera rig of PAI-AV~\citep{nvidia2025physicalai}, whose full surround views are absent from all perception training data. After correcting lens distortion to a pinhole model, we run \ours{}-SFT directly on both rigs without dataset-specific adaptation. As shown in Fig.~\ref{fig:ood_perception_vis}, the projected boxes are visually plausible across views and remain qualitatively consistent with the corresponding BEV layouts. The occupancy and map outputs also exhibit coherent road surfaces, object regions, drivable areas, lane markings, and road edges. These examples show that the perception pathway remains operational and produces qualitatively coherent outputs on unseen camera rigs. Because neither dataset provides unified ground truth, they do not establish reliable 3D accuracy under the new camera settings. Quantitative evaluation and adaptation with high-quality annotations from the target vehicle platform are required to assess and improve such cross-rig transfer.

\section{Conclusion}

We presented \ours{}, which we consider an initial step toward a vision-language foundation model for autonomous driving. The framework retains the pretrained VLM architecture and introduces external modules for unified 3D perception and motion planning. The BEV perception head serves as a 3D probe and adds explicit, inspectable detection, occupancy, and map predictions to the same pretrained VLM. The Planning Expert uses the shared representations to generate future ego trajectories through flow matching and supports joint training across multiple public driving datasets. A staged training and data recipe further combines driving-specific supervision with general-purpose vision-language data.

Experiments show that \ours{} achieves highly competitive results in 3D perception, driving scene understanding, and motion planning while largely preserving general vision-language capability. Its performance across open-loop, pseudo-closed-loop, and closed-loop planning evaluations also demonstrates the value of combining unified trajectory supervision with reward-based optimization. Together, these results show that explicit 3D perception and trajectory generation can be added to a pretrained VLM while retaining its broader vision-language competence.

\section{Limitations and Future Work}

Although \ours{} establishes a unified framework for perception, reasoning, and planning, several directions remain open. First, the planning reasoning does not always capture the causal structure of a scene accurately. Driving mixes causes that act at different time scales. A red light 20\,m ahead calls for early, gradual deceleration, whereas a child emerging 5\,m ahead demands an immediate response. When such causes coexist, the model remains unstable in identifying the governing cause and its temporal scope. Even when the suggested trend is appropriate, the decision executed within the next 1 to 2\,s may not reflect the stated immediate cause. Second, the generated trajectory does not always adhere to the textual rationale. Although reasoning improves downstream planning performance, part of this gain may stem from the additional model-internal information that the self-generated trace contributes to the conditioning context. Addressing these issues calls for multi-timescale causal modeling and explicit consistency supervision between the rationale and the generated trajectory.

Stronger cross-task transfer is another promising direction. The three tasks currently use different input formats, temporal contexts, and image resolutions, which may limit the transfer of learned representations. Better alignment of these configurations and closer joint optimization may allow perception, reasoning, and planning to reinforce one another more consistently.

\section*{Authors}

\bgroup

\textbf{Core Contributors:} 
Xin Zhou\textsuperscript{1,2},
Zongchuang Zhao\textsuperscript{1,2},
Zhibo Yang\textsuperscript{1}\textsuperscript{\,$\dagger$},
Mingsheng Li\textsuperscript{1},
Humen Zhong\textsuperscript{1},
Shuai Bai\textsuperscript{1},
Dingkang Liang\textsuperscript{2}\textsuperscript{\,\Letter},
Xiang Bai\textsuperscript{2}\textsuperscript{\,\Letter},
Dayiheng Liu\textsuperscript{1}\textsuperscript{\,$\dagger$}
\\[1.em]
\textbf{Contributors} (ordered alphabetically):
Du Chu\textsuperscript{1},
Ruizhe Chen\textsuperscript{1},
Zhaohai Li\textsuperscript{1},
Jun Tang\textsuperscript{1},
Qiuyue Wang\textsuperscript{1},
Mingkun Yang\textsuperscript{1},
Jiazhao Zhang\textsuperscript{1}
\\[1.em]
\textbf{External Advisors:} Dingkang Liang\textsuperscript{2}\textsuperscript{\,\Letter}, Xiang Bai\textsuperscript{2}\textsuperscript{\,\Letter}
\\[1.em]
\textsuperscript{1}\,Qwen Team \\
\textsuperscript{2}\,Huazhong University of Science and Technology
\\[0.5em]

{\let\thefootnote\relax\footnotetext{\hspace{-1.7em}\(\dagger\)~Project Leader. \quad \Letter~Corresponding authors.}

\textbf{Acknowledgment.} This work is partially supported by the NSFC (62225603).
 
 \egroup

\newpage
\bibliography{main}
\bibliographystyle{colm2024_conference}

\newpage
\appendix

\definecolor{okgreen}{HTML}{2E7D32}
\definecolor{badred}{HTML}{C62828}
\newcommand{\rok}{\textcolor{okgreen}{\cmark}}
\newcommand{\rbad}{\textcolor{badred}{\xmark}}

\section{Reward Definitions for Reinforcement Learning}
\label{app:rl_reward}

This appendix specifies the per-source rewards used in Stage 4. A rollout produces the trajectory $\boldsymbol{\tau}^{\mathrm{out}}=\{(\hat{x}_k,\hat{y}_k,\hat{\theta}_k)\}_{k=1}^{50}$ of Eq.~\ref{eq:traj}, covering 5\,s at 10\,Hz and expressed in metric units. Let $\boldsymbol{\tau}^{\mathrm{gt}}$ denote the recorded future trajectory. The displacement error over the first $n$ waypoints is:
\begin{equation}
\mathrm{ADE}_{n}\!\left(\boldsymbol{\tau}^{\mathrm{out}},\boldsymbol{\tau}^{\mathrm{gt}}\right)
=
\frac{1}{n}
\sum_{k=1}^{n}
\left\|
(\hat{x}_k,\hat{y}_k)
-
(x^{\mathrm{gt}}_k,y^{\mathrm{gt}}_k)
\right\|_2,
\label{eq:appade}
\end{equation}
which uses the positional channels only, excludes heading, and is measured in meters. Abbreviating it as $\mathrm{ADE}_{n}$, we write the shifted and scaled displacement term as:
\begin{equation}
\Delta(n;\delta,\kappa) = \frac{\delta - \mathrm{ADE}_{n}}{\kappa},
\label{eq:appshift}
\end{equation}
where $\delta$ centers the term and $\kappa$ sets its scale, both in meters. Every reward below is affine in the displacement errors. For a displacement term
$w\Delta(n;\delta,\kappa)$, $w$ denotes its weight. Because the group-relative advantage of Sec.~\ref{sec:training} standardizes rewards within a group, an additive constant and a common positive factor leave the policy gradient unchanged. The offsets $\delta$ therefore affect only the logged reward magnitude, while the ratios $w/\kappa$ determine the relative influence of the terms.

\paragraph{NAVSIM.} The reward combines PDMS with the full-horizon displacement term:
\begin{equation}
R_{\mathrm{NAVSIM}}
=
w_{\mathrm{pdms}}\,\mathrm{PDMS}
+
w_{\mathrm{ade}}\,\Delta(50;\delta,\kappa),
\label{eq:appnavsim}
\end{equation}
with $w_{\mathrm{pdms}}=1$, $w_{\mathrm{ade}}=2$, $\delta=2$, and $\kappa=10$. The NAVSIM PDM scorer evaluates the PDMS term against the scene metric cache. Each rollout is subsampled to the 8 poses at 2\,Hz over the 4\,s scoring horizon that the scorer expects, and the simulation interval is 0.1\,s. Within this reward, the sub-score weights of ego progress, time-to-collision, and comfort are 6, 4, and 2, whereas the evaluation reported in Sec.~\ref{sec:exp_planning} follows the official protocol.

\paragraph{WOD-E2E.} The reward combines the Rater Feedback Score with the same displacement term:
\begin{equation}
R_{\mathrm{WOD}}
=
w_{\mathrm{rfs}}\,\mathrm{RFS}
+
w_{\mathrm{ade}}\,\Delta(50;\delta,\kappa),
\label{eq:appwod}
\end{equation}
with $w_{\mathrm{rfs}}=1$, $w_{\mathrm{ade}}=2$, $\delta=2$, and $\kappa=1$. The RFS term is computed using the official rater-feedback utility against the scene's human preference trajectories and their rater scores. The rollout is linearly resampled to 20 positions at 4\,Hz over 5\,s to match the format the utility expects. Since RFS spans $[0,10]$ compared with $[0,1]$ for PDMS, we adopt a smaller $\kappa$ to keep the displacement term comparable in scale to the corresponding task reward.


\paragraph{PAI-AV.} PAI-AV defines no task-level score beyond displacement error, so its reward consists of displacement terms alone. Since near-term accuracy governs the immediate control decision, we combine the full horizon with shorter horizons under separate weights:
\begin{equation}
R_{\mathrm{PAI}}
=
w_{0}\,\Delta(50;\delta,\kappa_{0})
+
\sum_{h \in \{1,2,3,4\}}
w_{h}\,\Delta(10h;\delta,\kappa_{h}),
\label{eq:apppai}
\end{equation}
with $\delta=2$, $w_{0}=1$, $\kappa_{0}=1$, weights $(w_1,w_2,w_3,w_4)=(5,4,3,2)$, and scales $(\kappa_1,\kappa_2,\kappa_3,\kappa_4)=(2,4,6,8)$. The horizon $h$ corresponds to the first $10h$ waypoints. The resulting coefficients $w_h/\kappa_h$ are $2.5$, $1$, $0.5$, and $0.25$ for the 1\,s to 4\,s horizons against $1$ for the full 5\,s horizon, so an error in the first second carries the largest weight.

\paragraph{Summary.} Each source includes a displacement term. NAVSIM and WOD-E2E additionally use their task-level scores. The shared displacement term gives the three sources a comparable learning signal so that a single policy can be optimized across them. The multi-horizon refinement of Eq.~\ref{eq:apppai} is applied to PAI-AV only, since the task-level scores provide the dominant learning signal for the other two sources.

\section{Additional Visualizations}
\label{app:additional_vis}

Beyond the comparisons in Fig.~\ref{fig:drivevqa_vis}, this section illustrates a broader range of capabilities supported by \ours{}-SFT, including camera-based 3D grounding, traffic-signal detection, roadwork detection, road element recognition, and reasoning-based motion planning. The sequence progresses from structured visual prediction to traffic-rule interpretation and planning. We include prompts and raw model responses throughout. Where annotations are available, predictions from \ours{}-SFT and ground truth are overlaid on the input image in orange and blue, respectively.

\begingroup
\catcode`\"=\active
\def"{\textquotedbl}

\subsection{Camera-Based 3D Grounding}
\label{app:grounding_vis}

Camera-based 3D grounding requires recovering metric 3D boxes from a single image. In the urban turning scene below, \ours{}-SFT localizes three vehicles and reports their box parameters in the requested format.

\noindent\textbf{Prompt.}
\begin{quote}\small
Locate car in the provided image and output the corresponding 3D box. Camera intrinsics: fx=1862.42, fy=1861.99, cx=1917.51, cy=1078.37. Format: \texttt{[\{"bbox\_3d":[x\_center, y\_center, z\_center, x\_size, y\_size, z\_size, roll, pitch, yaw],"label":"category"\}]}.
\end{quote}

\begin{center}
\includegraphics[width=0.5\textwidth]{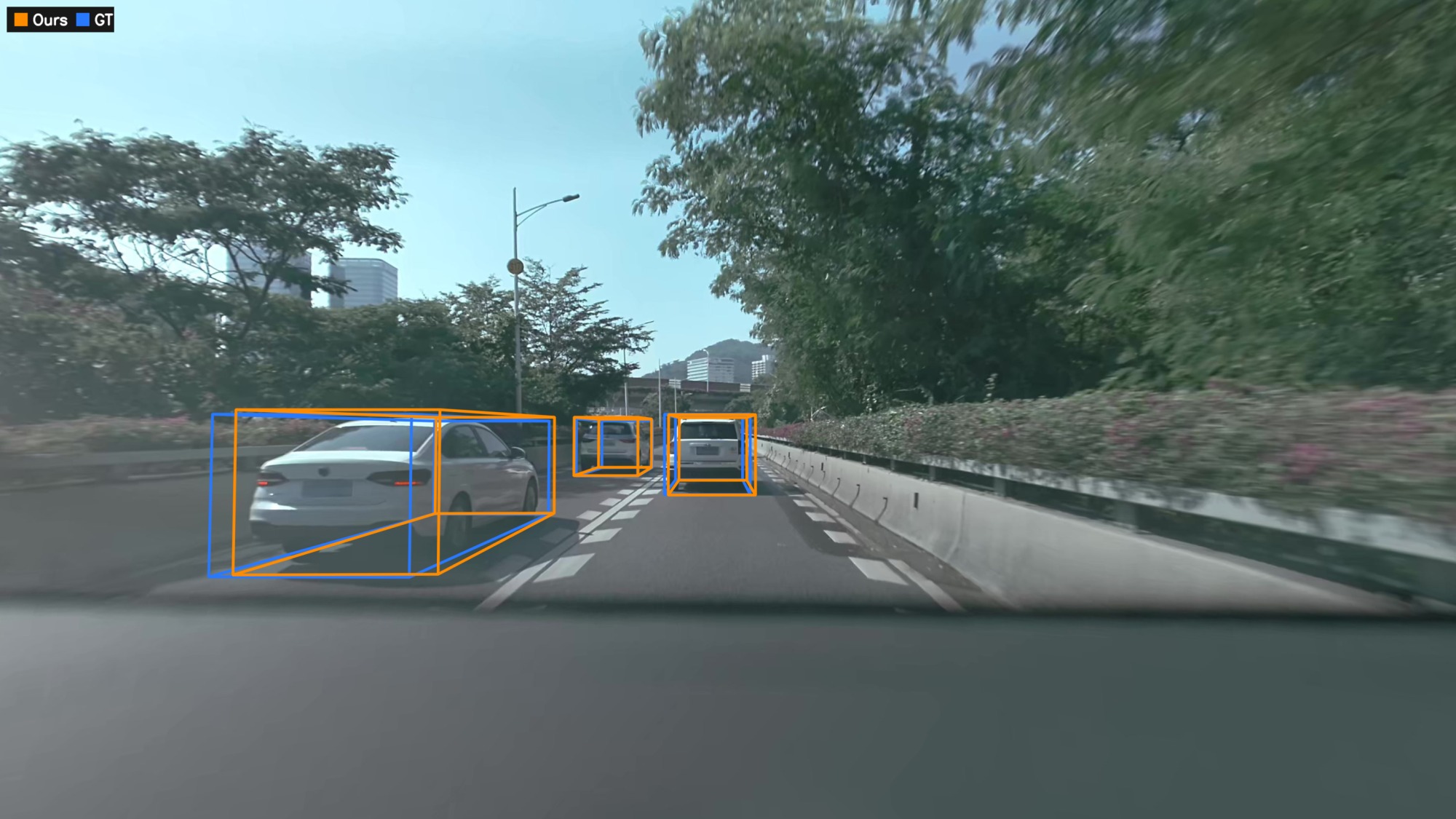}
\end{center}

\noindent\textbf{Response of \ours{}-SFT.}
\begin{quote}\small
\texttt{[\{"label": "car", "bbox\_3d": [-3.52, 0.80, 8.56, 4.51, 1.81, 1.47, -0.17, -0.49, 0.67]\},\\
\{"label": "car", "bbox\_3d": [-0.35, 1.01, 16.80, 4.22, 1.81, 1.67, -0.17, -0.49, 0.67]\},\\
\{"label": "car", "bbox\_3d": [-3.45, 1.11, 22.00, 4.22, 1.80, 1.65, -0.17, -0.49, 0.67]\}]}
\end{quote}

\subsection{Traffic-Signal Detection}
\label{app:traffic_light_vis}

At signalized intersections, detection must distinguish not only individual signals but also their type, orientation, and relevance to the current intersection. \ours{}-SFT localizes twelve signals in the frame and returns the requested attributes for each instance.

\noindent\textbf{Prompt.}
\begin{quote}\small
You are a visual annotation assistant for autonomous-driving intersection scenes. Given an image, detect and output all traffic-signal-related objects, focusing on the signal type (\texttt{label}), orientation (\texttt{direction}), and relevance to the current intersection (\texttt{is\_related}).

For every traffic-signal-related object in the image, output a dictionary containing:

-- \texttt{bbox\_2d}: the 2D bounding box in the format \texttt{[x1, y1, x2, y2]} using integer pixel coordinates. The top-left corner of the image is \texttt{(0,0)}, with \texttt{x1 < x2} and \texttt{y1 < y2}. Make the box as tight as possible around the visible signal body.

-- \texttt{label}: the signal type, one of \texttt{"car"} (motor-vehicle or non-motor-vehicle signal), \texttt{"pedestrian"} (pedestrian signal), \texttt{"black"} (black, unlit signal), \texttt{"digit"} (standalone countdown display), \texttt{"led"} (text LED sign), and \texttt{"other"} (another traffic signal or signal-control device).

-- \texttt{direction}: the signal orientation, one of \texttt{"front"} (facing the ego vehicle), \texttt{"back"} (facing away from the ego vehicle), and \texttt{"side"} (clearly facing sideways).

-- \texttt{is\_related}: whether the signal is relevant to the current intersection decision, one of \texttt{"related"} (controls traffic through the current intersection) and \texttt{"unrelated"} (a signal for a distant unrelated intersection, opposing traffic, or a side road).

Output exactly one JSON array of dictionaries with no extra explanatory text, wrapped in \texttt{```json ... ```}. If there are no traffic-signal-related objects, output an empty array. Output each object only once, and keep bounding-box coordinates as integers.
\end{quote}

\begin{center}
\includegraphics[width=0.5\textwidth]{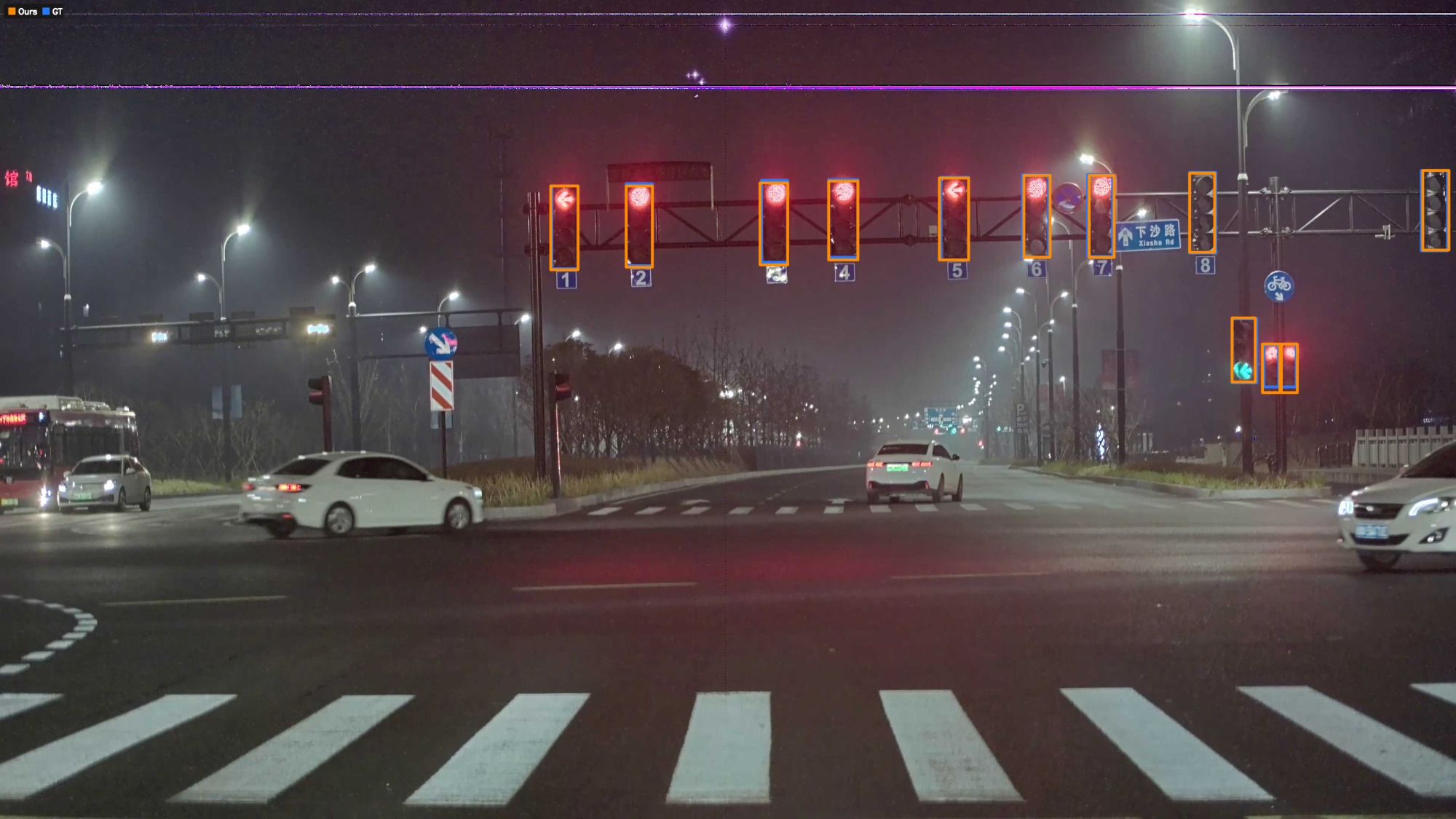}
\end{center}

\noindent\textbf{Response of \ours{}-SFT.}
\begin{quote}\small
\texttt{[\{"bbox\_2d":[378,227,397,329], "label":"car", "direction":"front", "is\_related":"related"\},\\
\{"bbox\_2d":[430,227,448,326], "label":"car", "direction":"front", "is\_related":"related"\},\\
\{"bbox\_2d":[522,223,541,323], "label":"car", "direction":"front", "is\_related":"related"\},\\
\{"bbox\_2d":[569,221,589,317], "label":"car", "direction":"front", "is\_related":"related"\},\\
\{"bbox\_2d":[645,217,665,317], "label":"car", "direction":"front", "is\_related":"related"\},\\
\{"bbox\_2d":[703,214,721,314], "label":"car", "direction":"front", "is\_related":"related"\},\\
\{"bbox\_2d":[748,214,765,314], "label":"car", "direction":"front", "is\_related":"related"\},\\
\{"bbox\_2d":[817,211,834,308], "label":"black", "direction":"front", "is\_related":"related"\},\\
\{"bbox\_2d":[977,208,995,305], "label":"black", "direction":"front", "is\_related":"related"\},\\
\{"bbox\_2d":[846,388,862,466], "label":"car", "direction":"front", "is\_related":"related"\},\\
\{"bbox\_2d":[867,420,879,479], "label":"car", "direction":"front", "is\_related":"related"\},\\
\{"bbox\_2d":[880,420,891,479], "label":"car", "direction":"front", "is\_related":"related"\}]}
\end{quote}

\subsection{Roadwork Detection}
\label{app:roadwork_vis}

Work zones broaden the object vocabulary to temporary traffic-control devices and construction-related objects. In the scene below, \ours{}-SFT identifies thirteen instances, including a work vehicle, tubular markers, and cones.

\noindent\textbf{Prompt.}
\begin{quote}\small
Locate every instance that belongs to the category set in the image. The category set includes: Police Officer, Police Vehicle, Cone, Fence, Drum, Barricade, Barrier, Work Vehicle, Vertical Panel, Tubular Marker, Arrow Board, Bike Lane, Work Equipment, Worker, Other Roadwork Objects, Temporary Traffic Control Message Board, Temporary Traffic Control Sign, Stop sign. Report bbox coordinates in JSON format like this: \texttt{[\{"bbox\_2d": [x1, y1, x2, y2], "label": "obj\_name"\}, ...]}.
\end{quote}

\begin{center}
\includegraphics[width=0.5\textwidth]{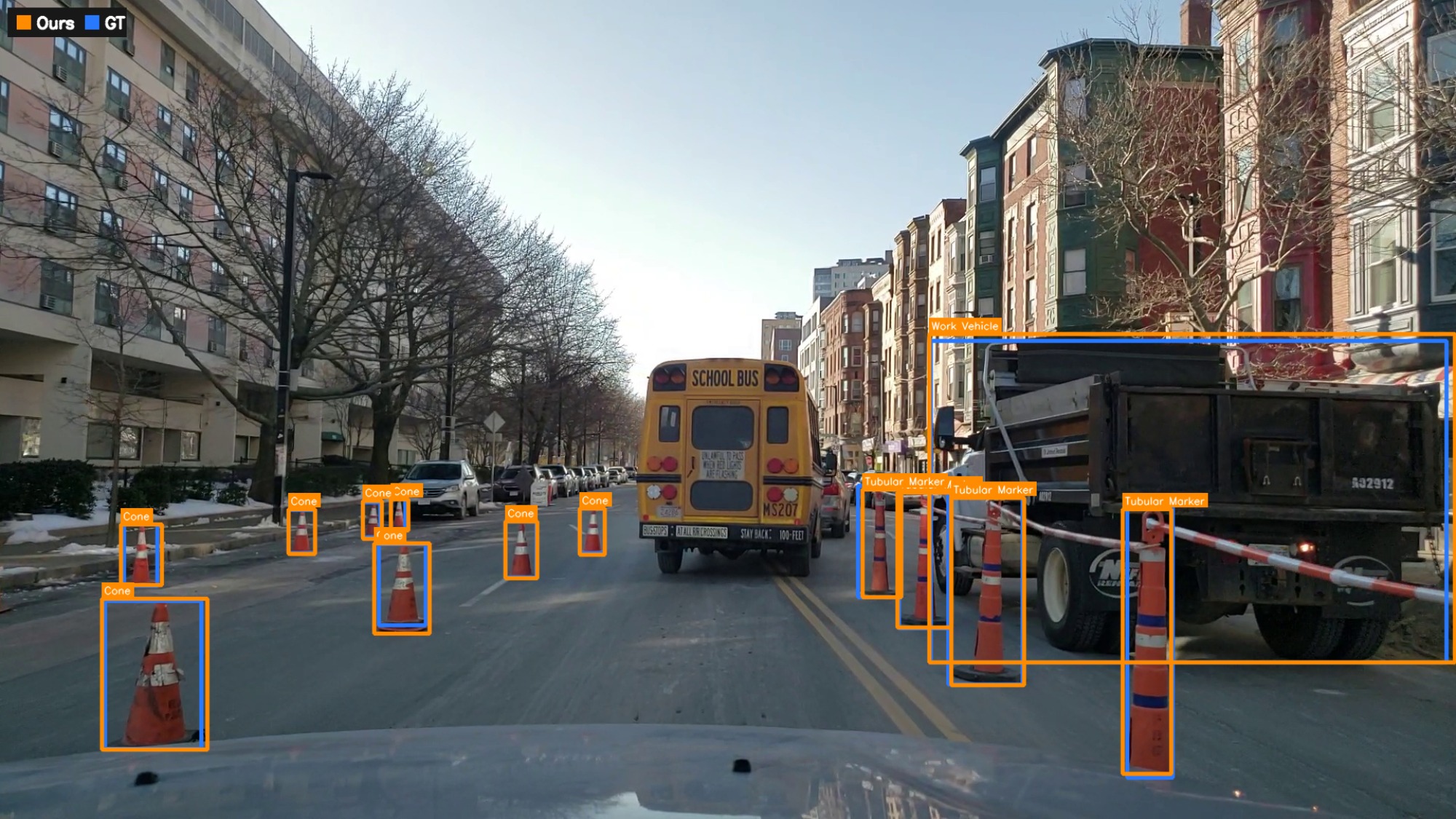}
\end{center}

\noindent\textbf{Response of \ours{}-SFT.}
\begin{quote}\small
\texttt{[\{"bbox\_2d":[638,407,999,808], "label":"Work Vehicle"\},\\
\{"bbox\_2d":[616,601,653,766], "label":"Tubular Marker"\},\\
\{"bbox\_2d":[592,597,619,729], "label":"Tubular Marker"\},\\
\{"bbox\_2d":[653,607,703,836], "label":"Tubular Marker"\},\\
\{"bbox\_2d":[771,621,804,944], "label":"Tubular Marker"\},\\
\{"bbox\_2d":[398,620,415,677], "label":"Cone"\},\\
\{"bbox\_2d":[347,636,369,706], "label":"Cone"\},\\
\{"bbox\_2d":[257,663,295,773], "label":"Cone"\},\\
\{"bbox\_2d":[268,609,280,647], "label":"Cone"\},\\
\{"bbox\_2d":[249,611,262,657], "label":"Cone"\},\\
\{"bbox\_2d":[198,621,216,676], "label":"Cone"\},\\
\{"bbox\_2d":[83,640,111,714], "label":"Cone"\},\\
\{"bbox\_2d":[70,731,142,915], "label":"Cone"\}]}
\end{quote}

\subsection{Road Element Recognition}
\label{app:road_element_vis}

Beyond object localization, \ours{}-SFT further interprets the traffic-rule semantics of road elements. Here, a red box highlights a single dashed yellow line on a Chinese expressway, whose meaning must be selected from six candidate answers. \ours{}-SFT selects the correct interpretation, while all eight comparison methods answer incorrectly.

\noindent\textbf{Prompt.}
\begin{quote}\small
The visual data is from China. What is the meaning of the Single Dashed Yellow Line within the red box in the image? Please select one phrase from the list below as the answer ['Vehicles must slow down and yield to main road vehicles or pedestrians', 'Straight and right-turn', 'Non-motorized vehicle lane', 'Located on both sides of the road, indicating a dedicated lane', 'Right-turn vehicle', 'Road leading to North of West Outer Ring']. Reason carefully and step-by-step to ensure logical accuracy and robustness, including any relevant error checks. Finally, provide the final answer within \texttt{\textbackslash boxed\{\}}.
\end{quote}

\begin{center}
\includegraphics[width=0.5\textwidth]{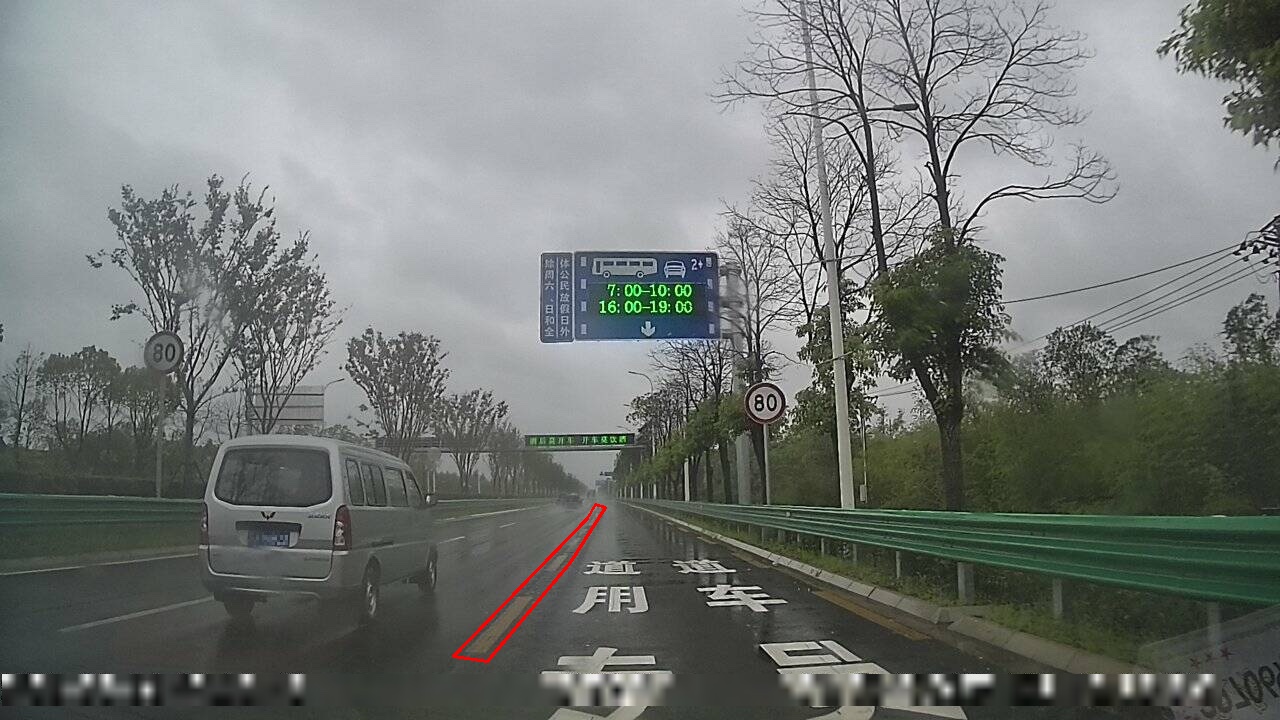}
\end{center}

\noindent\textbf{Response of \ours{}-SFT.}
\begin{quote}\small
To determine the meaning of the Single Dashed Yellow Line within the red box in the image, let us analyze the context and visual cues.

1. Visual context. The image is taken from a dashcam on a multi-lane highway or expressway in China (as indicated by Chinese characters on signs). The road surface has large white Chinese characters painted: "\begin{CJK}{UTF8}{gbsn}专用车道\end{CJK}", which translates to "Dedicated Lane". The red box highlights a single dashed yellow line, which typically separates lanes with different traffic rules or designations. In China, yellow lines are often used to indicate special-purpose lanes, such as bus lanes, emergency vehicle lanes, or dedicated turn lanes, especially when combined with other markings.

2. Road marking analysis. The word "\begin{CJK}{UTF8}{gbsn}专用\end{CJK}" means "dedicated" or "exclusive", and "\begin{CJK}{UTF8}{gbsn}车道\end{CJK}" means "lane", so "\begin{CJK}{UTF8}{gbsn}专用车道\end{CJK}" means "Dedicated Lane". The yellow line likely demarcates this dedicated lane from adjacent general-purpose lanes.

3. Evaluating the options. "Vehicles must slow down and yield to main road vehicles or pedestrians" describes a yield line or stop line, not a lane divider, and is incorrect. "Straight and right-turn" refers to directional arrows, not a yellow line, and is incorrect. "Non-motorized vehicle lane" is incorrect, because non-motorized vehicles usually have separate bike lanes often marked with blue or green paint and bicycle symbols, and yellow lines are rarely used for non-motorized lanes on highways. "Located on both sides of the road, indicating a dedicated lane" matches the analysis: the yellow line is part of a marking system that designates a dedicated lane ("\begin{CJK}{UTF8}{gbsn}专用车道\end{CJK}"). Although the phrase says "on both sides", the key part is "indicating a dedicated lane", which aligns with the road marking text. "Right-turn vehicle" again refers to directionality rather than lane designation via a yellow line, and is incorrect. "Road leading to North of West Outer Ring" is a geographic description unrelated to lane markings, and is incorrect.

4. Conclusion. The only option that correctly interprets the purpose of the single dashed yellow line in context with the surrounding road markings ("\begin{CJK}{UTF8}{gbsn}专用车道\end{CJK}") is "Located on both sides of the road, indicating a dedicated lane". Even though "both sides" may be slightly misleading (since we only see one side), the core meaning, indicating a dedicated lane, is accurate and matches the Chinese text on the road.

Final Answer: \texttt{\textbackslash boxed\{Located on both sides of the road, indicating a dedicated lane\}}
\end{quote}

\subsection{Reasoning-Based Motion Planning}
\label{app:navsim_coc_vis}

Finally, we turn from scene interpretation to future-motion prediction. \ours{}-SFT supports predicting a trajectory with a one-sentence rationale. Given multiview temporal observations, the historical ego trajectory, and the active navigation command, \ours{}-SFT produces a one-sentence rationale followed by future waypoints serialized as text. The ego vehicle follows the lead vehicle through a signalized intersection, and the predicted trajectory closely tracks the ground truth in both the camera view and the BEV visualization.

\noindent\textbf{Input.}
\begin{center}
\includegraphics[width=\textwidth]{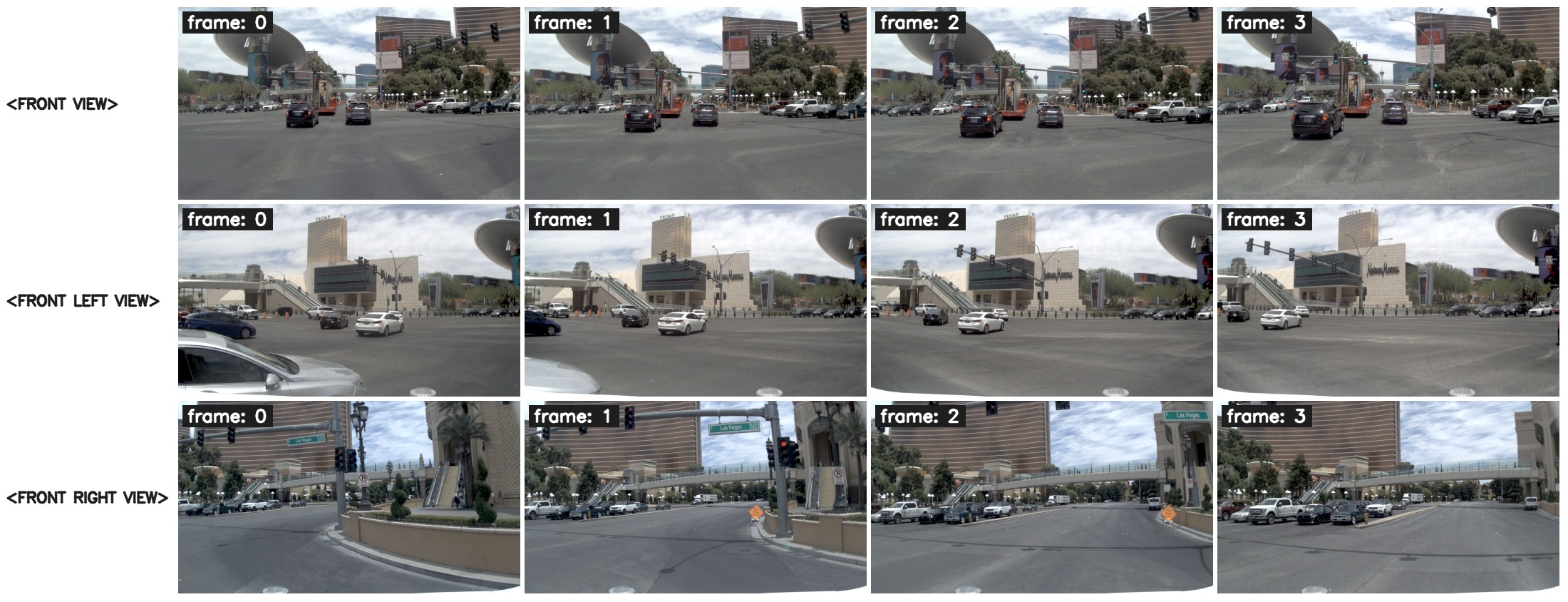}
\end{center}

\noindent\textbf{Prompt.}
\begin{quote}\small
The input images are organized by camera view. Each view contains 4 temporal frames captured at 0.5s intervals (frame 0 at t-1.5s, frame 3 is the current frame at t=0s).

1. Historical trajectories (x, y, heading) in the current frame's ego coordinate system. Positive x points forward, positive y points left, and a positive heading indicates a left turn:

-t-1.5s: (-15.6548, 0.0502, 0.0069)\\
-t-1.0s: (-10.3809, 0.0326, 0.0042)\\
-t-0.5s: (-5.1587, 0.0116, 0.0023)\\
-t-0s: (0.0000, 0.0000, 0.0000)

2. Active navigation command: [GO STRAIGHT]

First give a one-sentence brief reasoning of the ego's future driving decision, then predict the corresponding possible trajectory (2 Hz, 4 s, 8 points, each point (x, y, heading) with 2-decimal precision). Report in JSON format like this: \texttt{[\{"reasoning": "<one concise sentence>"\}, \{"trajectory": "[PT, (x, y, heading), ...]"\}]}.
\end{quote}

\noindent\textbf{Response of \ours{}-SFT.}
\begin{quote}\small
\texttt{[\{"reasoning": "Follow the lead vehicle through the intersection under the green traffic light."\},\\
\{"trajectory": "[PT, (5.09, -0.01, 0.00), (10.12, -0.02, 0.00), (15.10, -0.03, 0.00), (20.03, -0.04, 0.00), (24.91, -0.04, 0.00), (29.74, -0.03, 0.00), (34.52, -0.02, 0.00), (39.24, -0.01, 0.00)]"\}]}
\end{quote}

\begin{center}
\includegraphics[width=0.75\textwidth]{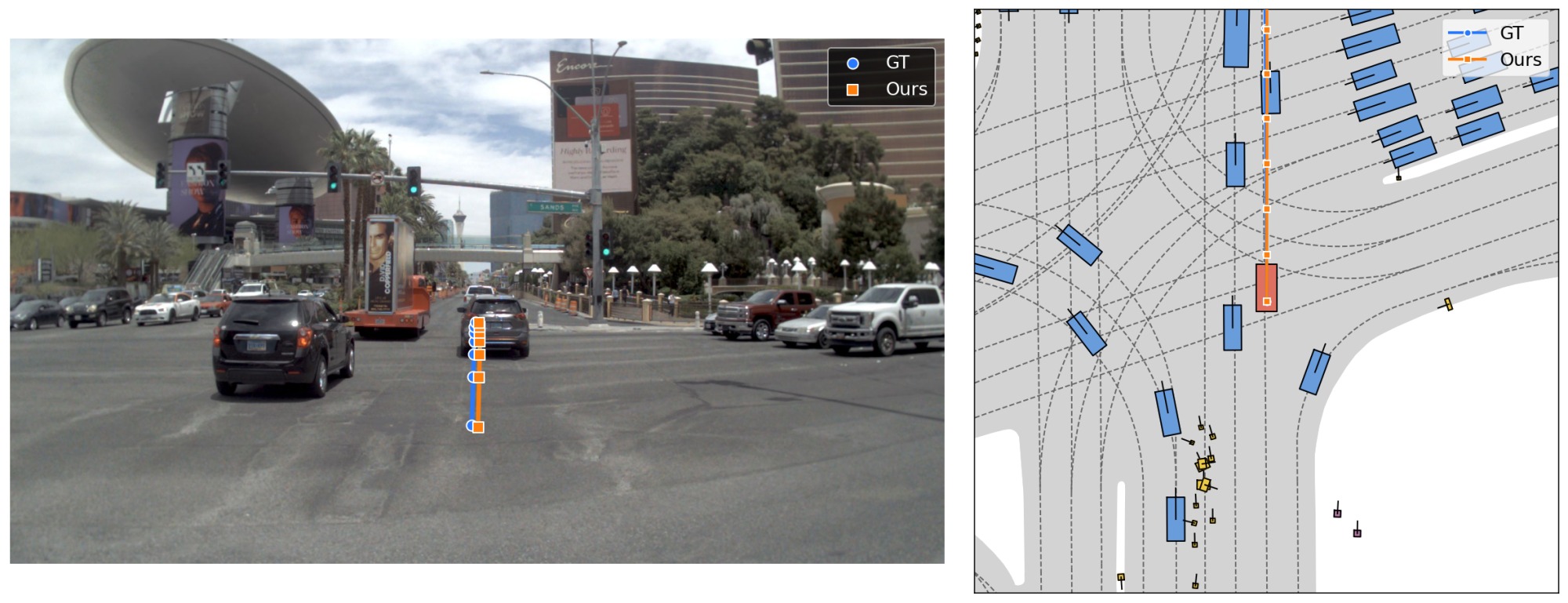}
\end{center}

\endgroup

\section{Detailed Driving VQA Cases}
\label{app:drivevqa_cases}

This appendix reports the complete prompts, ground-truth answers, and model responses for the four cases summarized in Fig.~\ref{fig:drivevqa_vis}. Every model receives the same images in the same order. Two edits are applied to the raw outputs: markdown emphasis is typeset as text, and long chains of thought are abridged at positions marked with [\dots], where the omitted spans restate the surrounding reasoning without adding evidence.

\subsection{(a) Temporal Understanding and Agent-State Estimation (LingoQA)}

\begin{quote}\small
What you are seeing are frames from a video sequence. Your answer should focus on the final dynamic outcome, rather than analyzing the results of each individual frame.

How many parked vehicles can you see?
\end{quote}

\paragraph{Ground truth.} \emph{Zero.} The benchmark also accepts \emph{None.}

\noindent\textbf{Responses.}

\noindent\rok~\textbf{\ours{}-SFT}
\begin{quote}\small
There are no vehicles parked on either side of the road.
\end{quote}

\noindent\rbad~\textbf{Qwen3.5-4B}
\begin{quote}\small
In the provided video frames, I can see the following parked vehicles:

-- A red bus is parked on the left side of the road.\\
-- A white truck is parked on the right side of the road.

 Therefore, there are \textbf{2 parked vehicles} visible in the video sequence.
\end{quote}

\noindent\rbad~\textbf{MiMo-Embodied-7B}
\begin{quote}\small
\texttt{\textless think\textgreater}\\
So, let's look at the frames. In frame 5, there are parked vehicles on the right side. Let's count: there's a white van, a car, and maybe others? Wait, no, let's check. [\dots\ the trace repeats this self-correction for roughly four thousand characters \dots] Wait, the correct answer is three.\\
\texttt{\textless/think\textgreater}

To determine the number of parked vehicles, we analyze the frames. In the final frame (frame 5), we observe parked vehicles on the right-hand side of the road. Counting these, we can identify three parked vehicles: a white van, a car, and another vehicle.

So, the number of parked vehicles is 3.
\end{quote}

\subsection{(b) Causal Reasoning for Planning (PAI-AV-CoC)}

\begin{quote}\small
The input images are organized by camera view. Each view contains 4 temporal frames captured at 0.5s intervals (frame 0 at t-1.5s, frame 1 at t-1.0s, frame 2 at t-0.5s, frame 3 is the current frame at t=0s).

1. Historical trajectories (x, y, heading) in the current frame's ego coordinate system. Positive x points forward, positive y points left, and a positive heading indicates a left turn:

-t-1.5s: (-13.9670, -0.4070, 0.0427); -t-1.0s: (-8.9300, -0.1980, 0.0370); -t-0.5s: (-4.2570, -0.0560, 0.0219); -t-0s: (0.0000, 0.0000, 0.0000);

Give a one-sentence brief reasoning of the ego's future driving decision ONLY.
\end{quote}

\paragraph{Ground truth.} \emph{Stop behind the lead vehicle at the stop sign at the intersection.}

\noindent\textbf{Responses.}

\noindent\rok~\textbf{\ours{}-SFT}
\begin{quote}\small
Stop for the stop sign at the intersection.
\end{quote}

\noindent\rbad~\textbf{Cosmos-Reason2-32B}
\begin{quote}\small
The ego vehicle will continue straight, maintaining its lane position while navigating around the construction barriers on the right side of the road.
\end{quote}

\noindent\rbad~\textbf{Alpamayo-1.5-10B}
\begin{quote}\small
The ego vehicle is going straight at a steady speed
\end{quote}

\subsection{(c) Cross-View Spatial Distance Perception (Ego3D-Bench)}

\begin{quote}\small
These are seven camera views mounted on an ego car. Front Left view: \texttt{\textless image\textgreater}; Front view: \texttt{\textless image\textgreater}; Front Right view: \texttt{\textless image\textgreater}; Right view: \texttt{\textless image\textgreater}; Back Right view: \texttt{\textless image\textgreater}; Back Left view: \texttt{\textless image\textgreater}; Left view: \texttt{\textless image\textgreater}.

How far, in meters, is the dark colored sedan car parked on the side of the street in the back right view from the beige sedan car parked on the side of the street in the front view?

Output the thinking process and final answer (number ONLY) in \texttt{\textbackslash boxed\{\}} tags.
\end{quote}

\paragraph{Ground truth.} 22.93\,m.

\noindent\textbf{Responses.}

\noindent\rok~\textbf{\ours{}-SFT}
\begin{quote}\small
To determine the distance between the two cars, we need to analyze their positions in the provided views.

1. \textbf{Identify the beige sedan car in the front view:}\\
-- In the \texttt{\textless FRONT VIEW\textgreater}, there is a beige sedan parked on the right side of the street. It is positioned behind a white SUV and in front of a silver sedan (which is partially visible). The beige sedan is clearly visible with its rear facing the camera.

2. \textbf{Identify the dark colored sedan car in the back right view:}\\
-- In the \texttt{\textless BACK RIGHT VIEW\textgreater}, there is a dark gray or black Mercedes-Benz sedan parked on the left side of the street (from the perspective of the back right view). This car is parked on the curb next to some bushes and trees.

3. \textbf{Estimate the distance:}\\
-- The two cars are on opposite sides of the street. The street appears to be a typical urban road with a width of about 6-8 meters (including the parking lanes).\\
-- The beige sedan is parked on the right side, and the dark sedan is parked on the left side.\\
-- Additionally, there is a significant distance along the street between the two cars. In the \texttt{\textless FRONT VIEW\textgreater}, the beige sedan is relatively close to the camera. In the \texttt{\textless BACK RIGHT VIEW\textgreater}, the dark sedan is further down the street, past several parked cars and trees.\\
-- Based on the visual spacing and the typical dimensions of such streets, the distance between the two cars can be estimated. The width of the street alone adds about 6-8 meters. The longitudinal distance along the street appears to be at least 10-15 meters based on the perspective and the number of parked cars between them.

4. \textbf{Calculate the total distance:}\\
-- Combining the transverse distance (width of the street) and the longitudinal distance (distance along the street), the total distance is approximately 20-25 meters.

Given the estimation, the distance is around 20-25 meters.

\texttt{\textbackslash boxed\{22\}}
\end{quote}

\noindent\rbad~\textbf{Cosmos-Reason2-32B}
\begin{quote}\small
Okay, let's try to figure out the distance between the dark-colored sedan in the back right view and the beige sedan in the front view.

[\dots] Since both are on the same street but in different parts (front and back), they might be along the same road but separated by some distance. However, the problem doesn't provide explicit measurements. [\dots]

But I'm not sure. Maybe the answer is 10 meters. Alternatively, maybe the answer is that they are parked in different parts of the street, so the distance is more than 10 meters.

Alternatively, perhaps the answer is that they are parked on the same side of the street but separated by a distance of approximately 15 meters.

Given that, I'll go with 15 meters as the estimated distance.\\
\texttt{\textless/think\textgreater}

\texttt{\textbackslash boxed\{15\}}
\end{quote}

\noindent\rbad~\textbf{UniDriveVLA-8B}
\begin{quote}\small
The dark colored sedan car parked on the side of the street in the back right view is approximately 15 meters away from the beige sedan car parked on the side of the street in the front view. This estimation is based on the relative positions of the cars in the images, where the dark sedan is visible further down the street in the back right view, indicating a significant distance from the beige sedan in the front view. The visual evidence from the images supports the original GPT answer, confirming the distance as approximately 15 meters.
\end{quote}

\subsection{(d) Traffic-Road Recognition (VLADBench)}

\begin{quote}\small
The following 5 frames are from China in chronological order. The visual data is from China. In the given autonomous driving image sequence, at the moment of the final image, what is the ego lane? Please choose the most appropriate answer from the following options: ['straight lane', 'left turn lane', 'right turn lane', 'straight and left turn lane', 'straight and right turn lane', 'U-turn lane', 'left U-turn lane']. Reason carefully and step-by-step to ensure logical accuracy and robustness, including any relevant error checks. Finally, provide the final answer within \texttt{\textbackslash boxed\{\}}.
\end{quote}

\paragraph{Ground truth.} \emph{left turn lane}

\noindent\textbf{Responses.}

\noindent\rok~\textbf{\ours{}-SFT}
\begin{quote}\small
To determine the ego lane at the moment of the final image (Frame 5), we need to analyze the road markings visible in the sequence, particularly in the earlier frames where they are clearer.

1. \textbf{Observe Road Markings}: In Frame 1 and Frame 2, the road surface has clear directional arrows painted on it.\\
-- The lane the ego vehicle is currently in has a large arrow indicating a \textbf{left turn}.\\
-- The lane to the right of the ego vehicle (where the black SUV and bus are) has an arrow indicating a \textbf{right turn}.

2. \textbf{Track Vehicle Position}: Throughout the sequence from Frame 1 to Frame 5, the ego vehicle remains in the same lane --- the one with the left-turn arrow. It does not change lanes.

3. \textbf{Traffic Light Context}: The traffic light ahead is red, so all vehicles are stopped or slowing down. This confirms that the ego vehicle is waiting at an intersection in a designated turning lane.

4. \textbf{Eliminate Other Options}:\\
-- ``Straight lane'' --- incorrect, because the arrow is for turning.\\
-- ``Right turn lane'' --- that's the adjacent lane.\\
-- ``Straight and left/right turn lane'' --- no such combined marking is visible; each lane has a single directional arrow.\\
-- ``U-turn lane'' --- not indicated by the arrow shape (it's a standard left-turn arrow, not a U-turn symbol).

Therefore, based on the persistent left-turn arrow in the ego vehicle's lane throughout the sequence, especially clear in Frame 1 and 2, the ego lane is the \textbf{left turn lane}.

\texttt{\textbackslash boxed\{left turn lane\}}
\end{quote}

\noindent\rbad~\textbf{UniDriveVLA-8B}
\begin{quote}\small
The ego lane is the straight lane. This is evident from the sequence of images where the vehicle maintains a consistent position in the lane without any indication of turning or changing lanes. The lane markings and the surrounding traffic flow also support this conclusion, as there are no signs of a turn or lane change in the final image. Therefore, the most appropriate answer is 'straight lane'.

\texttt{\textbackslash boxed\{straight lane\}}
\end{quote}

\noindent\rbad~\textbf{Alpamayo-1.5-10B}
\begin{quote}\small
The most appropriate answer is \texttt{\textless c1,CAM\_FRONT,970.0,511.7\textgreater}.
\end{quote}

\end{document}